\documentclass[twoside,11pt]{article} 
\usepackage{jmlr2e} 

\usepackage{amsmath} 
\usepackage{amsxtra} 
\usepackage{amssymb} 
\usepackage{latexsym} 
\usepackage[titles]{tocloft} 
\usepackage{algorithm}
\usepackage{algorithmic}

\usepackage{lscape}
\usepackage{float}
\usepackage[hang,multiple]{footmisc}

\newif\iffootpunct
\footpuncttrue

\firstpageno{1}

\begin{document}
\title{Hybridized Threshold Clustering for Massive Data}
\author{\name Jianmei Luo \thanks{Some of the computing for this project was performed on the Beocat Research Cluster at Kansas State University, which is funded in part by NSF grants CNS-1006860, EPS-1006860, EPS-0919443, ACI-1440548, CHE-1726332, and NIH P20GM113109.} \email jianmei@k-state.edu \\
      \addr Department of Statistics\\
      Kansas State University\\
      Manhattan, KS 66506-0802, USA
        \AND
        \name ChandraVyas Annakula \email chandraa@k-state.edu \\
        \addr Department of Computer Science\\
        Kansas State University\\
        Manhattan, KS 66506-0802, USA
        \AND
        \name Aruna Sai Kannamareddy \email arunasai@k-state.edu\\
        \addr Department of Computer Science\\
        Kansas State University\\
        Manhattan, KS 66506-0802, USA        
        \AND
        \name Jasjeet S. Sekhon \thanks{The author would like to thank Office of Naval Research (ONR) Grant N00014-17-1-2176}\email sekhon@berkeley.edu \\
        \addr Department of Political and Statistics\\ University of California, Berkeley\\ Berkeley, CA 94720-1950, USA
        \AND
        \name William Henry Hsu \thanks{This work was supported in part by the Laboratory Directed Research and Development (LDRD) program at Lawrence Livermore National Laboratory (16-ERD-019). Lawrence Livermore National Laboratory is operated by Lawrence Livermore National Security, LLC, for the U.S. Department of Energy, National Nuclear Security Administration under Contract DE-AC52-07NA27344.} \email bhsu@k-state.edu \\
        \addr Department of Computer Science\\
        Kansas State University\\
        Manhattan, KS 66506-0802, USA
        \AND
        \name Michael Higgins \email mikehiggins@k-state.edu \\
      \addr Department of Statistics\\
      Kansas State University\\
      Manhattan, KS 66506-0802, USA}


 \maketitle
\begin{abstract}
As the size $n$ of datasets become massive, many commonly-used clustering algorithms (for example, $k$-means or hierarchical agglomerative clustering (HAC) require prohibitive computational cost and memory.
In this paper, we propose a solution to these clustering problems by extending threshold clustering (TC) to problems of instance selection.
TC is a recently developed clustering algorithm designed to partition data into many small clusters in linearithmic time (on average). 
Our proposed clustering method is as follows.
First, TC is performed and clusters are reduced into single ``prototype'' points.
Then, TC is applied repeatedly on these prototype points until sufficient data reduction has been obtained.  
Finally, a more sophisticated clustering algorithm is applied to the reduced prototype points, thereby obtaining a clustering on all $n$ data points.
This entire procedure for clustering is called \textit{iterative hybridized threshold clustering} (IHTC). 
Through simulation results and by applying our methodology on several real datasets, we show that IHTC combined with $k$-means or HAC substantially reduces the run time and memory usage of the original clustering algorithms while still preserving their performance.
Additionally, IHTC helps prevent singular data points from being overfit by clustering algorithms.

\end{abstract}

\begin{keywords}
  Threshold Clustering, Hybridized Clustering, Instance Selection, Prototypes, Massive Data
\end{keywords}

\section{Introduction}
\label{makereference1.1}
Clustering, also known as unsupervised learning, is a well-studied problem in machine learning. 
It aims to group units with similar features together and separate units with dissimilar features \citep{friedman2001elements}.
Cluster analysis has been used in many fields like biology, management, pattern recognition, etc. 
Additionally, many methods (for example, $k$-means clustering, hierarchical agglomerative clustering, etc.) have been developed that successfully tackle the clustering problem. 

However, enormous amounts of data are collected every day. 
For example, Walmart performs more than 1 million customer transactions per hour~\citep{cukier2010data}, and Google performs more than 3 billion searches per day~\citep{sullivan2015google}. This becomes a massive accumulation of data.
When working with data of such a large \textit{size}, 
many of the state-of-the-art clustering methods become intractable. 
That is, massive data requires novel statistical methods to process this data; research on scaling up existing statistical algorithms and scaling down the size of data without loss of information is of critical importance~\citep{jordan2015machine}.

Instance selection is a commonly-used pre-possessing method for scaling down the size of data~\citep{liu1998feature,blum1997selection,liu2002issues}.
The goal of instance selection is to shorten the execution time of data analysis by reducing data size $n$ while maintaining the integrity of data \citep{olvera2010review}. 
Instance selection methods rely on sampling, classification algorithms, or clustering algorithms. 
Previous work has shown methods reliant on clustering have better performance (accuracy) than some methods that rely on classification \citep{riquelme2003finding, raicharoen2005divide, olvera2007object, olvera2008prototype, olvera2010review}. 
However, current instance selection methods that rely on  classification often have faster runtimes.

On the other hand, threshold clustering (TC) is a recently developed method for clustering that is extremely efficient.
TC is a method of clustering units so that each cluster contains at least a pre-specified number of units $t^*$ while ensuring that the within-cluster dissimilarities are small. 
Previous work has shown that, when the objective is to minimize the maximum within-cluster dissimilarities, a solution within a factor of four of optimal can often be obtained in $O(t^*n)$ time and space when the $(t^*-1)$-nearest-neighbors graph is given~\citep{higgins2016improving}. 
The runtime and memory usage required for TC is smaller compared to other clustering methods, for example, $k$-means and hierarchical agglomerative clustering (HAC).

In this paper, we propose the use of TC for instance selection. 
The proposed method, which is called iterated threshold instance selection (ITIS), works as follows. 
For a given $t^*$, TC is applied on $n$ units to form $n^*$ clusters; each cluster will contain at least $t^*$ units. 
Then prototypes are formed by finding the center of each cluster. 
TC is applied again to the $n^*$ prototypes if the data is not sufficiently reduced. 
Otherwise, the procedure is stopped.

We also propose using ITIS as a pre-processing step on large data to allow for the use of more sophisticated clustering methods.
First, ITIS is applied to form a sufficiently small set of prototypes.  
Then, a more sophisticated clustering algorithm (for example $k$-means, HAC) is applied on this set of prototypes.
Finally, a clustering on all $n$ units is obtained by "backing out"  the cluster assignments for the prototypes---for each prototype, the units used to form the prototype are determined; these units are assigned to the same cluster assigned to the prototype. 
This clustering process on all $n$ units is called Iterative Hybridized Threshold Clustering (IHTC). 

We show, using simulations and applications of our algorithm to six large datasets, that IHTC combined with other clustering algorithms reduces the run time and memory usage of the original clustering algorithms while still preserving their performance.
Additionally, we show IHTC also prevents singular data points from being overfit by desired clustering methods. 
Specifically, for $m$ iterations of ITIS at size threshold $t^*$, IHTC ensures that each cluster contains at least $(t^*)^m$ units.

The rest of this paper is organized as follows. 
A brief summary about clustering algorithms ($k$-means, HAC, TC) is given in section \ref{makereference1.2}. 
Section \ref{makereference1.3} shows how to extend threshold clustering as an instance selection method and combine the iterated threshold instance selection method with other clustering methods. 
A simulation study is presented in section \ref{makereference1.4} and application of our methods on real datasets are presented in section \ref{makereference1.5}. 
The last section discuss about our method.

\section{Notation and Preliminaries}
\label{makereference1.2}
Consider a dataset with n units, numbered 1 through $n$. 
Each unit $i$ has a response vector $\mathbf y_i$ and a $d$-dimensional covariate vector $\mathbf x_i=(x_{i1},x_{i2},\ldots, x_{id})$. 
For each pair of units $i,j$, the \textit{dissimilarity}  between $i$ and $j$, denoted $d_{ij}$, can be computed.  
Often, the dissimilarity is chosen so that, if units $i$ and $j$ have similar values of covariates $\mathbf x$, then $d_{ij}$ is small. 
We assume that $d_{ij} \geq 0$, and that dissimilarities  satisfy the triangle inequality; for any units $i$, $j$, $\ell$,
\begin{equation}
    \label{eq:triangleineq}
    d_{ij} + d_{jk} \geq d_{i\ell}
\end{equation}
Common choices of dissimilarities include Euclidean distance, Manhattan distance and average distance.

We define a \textit{clustering} of a set of units as a partitioning of units such that units within each cluster of a partition have ``small dissimilarity'' and units belonging to two different clusters have ``large dissimilarity.''
That is, at minimum, a clustering $\mathbf v = \{V_1, V_2,\ldots, V_m\}$ will satisfy the following properties:
\begin{enumerate}
    \setlength{\itemsep}{1pt}
	\setlength{\parskip}{0pt}
	\setlength{\parsep}{0pt}
    \item \textbf{(Non-empty)}: $V_\ell \neq \emptyset$ for all $V_\ell \in \mathbf v$.
    \item \textbf{(Spanning)}: For all units $i$, there exists a cluster $V_\ell \in \mathbf v$ such that $i \in V_\ell$.
    \item \textbf{(Disjoint)}: For any two clusters $V_\ell, V_{\ell'} \in \mathbf v$, $V_\ell \cap V_{\ell'} = \emptyset$
\end{enumerate}
The way of measuring ``large'' and ``small'' cluster dissimilarity will vary across clustering algorithms.

There are currently hundreds of available methods for clustering units.
Moreover, some of these methods may be combined to construct additional \textit{hybridized} clustering methods---our procedure for hybridizing is the major contribution of this paper.
For brevity, we apply our hybridizing procedure to two clustering methods---$k$-means and hierarchical clustering---with a note that this procedure may be applied to many other types of clustering.
We now give a brief summary of these clustering methods.

\subsection{K-means Clustering}
\label{makereference1.2.1}
The $k$-means clustering algorithm \citep{lloyd1982least} is one of the most widely used and effective methods that attempts to partition units into exactly $k$-clusters.  

The $k$-means clustering algorithm proceeds as follows:
\begin{enumerate}
	\setlength{\itemsep}{1pt}
	\setlength{\parskip}{0pt}
	\setlength{\parsep}{0pt}
	\item \label{km_step1} 
	\textbf{(Initialization)} 
	Randomly select a set of $k$ units (referred as \textit{centers}) from the dataset. $K$ denote the number of clusters and it should be pre-specified. 
	\item \label{km_step2} 
	\textbf{(Assignment)} Assign all the units to the nearest center, based on squared Euclidean distance, to form $k$ temporary clusters.
    \item \label{km_step3} \textbf{(Updating)} Recompute the mean of each cluster. 
    Replace the \textit{centers} with the new $k$ cluster means.
	\item \label{km_step4}\textbf{(Terminate)} Repeat step~\ref{km_step2} and ~\ref{km_step3} until there is no further change for the \textit{centers}.
\end{enumerate}
The time complexity for the $k$-means clustering algorithm is $O(nkLd)$ and the space complexity is $O((k+n)d)$~\citep{hartigan1979algorithm,firdaus2015survey} where $d$ is the number of attributes for each unit, $L$ is the number of iterations taken by the algorithm to converge. 

The $k$-means algorithm suffers from a number of drawbacks ~\citep{hastie2009unsupervised}.
First, there tends to be high sensitivity to the selection of initial units in Step~\ref{km_step1}.
Additionally, it tends to overfit isolated units leading to some clusters containing only a few units.
Finally, the number of clusters $k$ is fixed; if $k$ is misspecified, $k$-means may perform poorly. 
In particular, many methods have been developed to mitigate problems due to initialization \citep{franti1997genetic, frnti1998tabu, arthur2007k, franti2000randomised}.



\subsection{Hierarchical Agglomerative Clustering}
\label{makereference1.2.2}
Hierarchical agglomerative clustering (HAC) \citep{ward1963hierarchical} is a ''bottom up''
approach that aims to build a hierarchy clusters. 
It initially treats each unit as a cluster and then continues to merge two clusters together until only one cluster remains.
HAC does not require a pre-specified number of clusters; the desired number of clusters can be obtained by using a \textit{dendogram}---a tree that shows how the units are merged.

The HAC proceeds as follows:
\begin{enumerate}
	\setlength{\itemsep}{1pt}
	\setlength{\parskip}{0pt}
	\setlength{\parsep}{0pt}
	\item \label{hac_step1} \textbf{(Initial Clusters)} Start with $n$ clusters, each cluster only contains one unit.
	\item \label{hac_step2} \textbf{(Merge)} Merge the closest (most similar) pair of clusters into a single cluster.
    \item \label{hac_step3} \textbf{(Updating)} Recompute the distance between the new cluster and the original clusters.
	\item \label{hac_step4}\textbf{(Terminate)} Repeat step~\ref{hac_step2} and ~\ref{hac_step3} until one cluster remains, the cluster contains $n$ units. 
\end{enumerate}

HAC requires linkage criteria to measure inter-cluster distance, but initialization and the choice of $k$ is no longer a problem. 
However, the time complexity of HAC is $O(n^{2}\log(n))$ \citep{kurita1991efficient} and space complexity is $O(n^2)$ \citep{jain1999data}. 
This complexity limits its application to massive data.
Another hindrance of HAC is that every merging decision is final. 
Once two clusters are merged into a new cluster, there is no way to partition the new cluster in later steps. 

\subsection{Threshold Clustering (TC)}
\label{makereference1.2.3}

    
Our hybridization method makes use of a recently developed clustering method called \textit{threshold clustering} (TC).
Initially this method was developed for performing statistical blocking of massive experiments~\citep{higgins2016improving}.
TC differs in two significant ways from traditional clustering approaches. 
First, TC does not fix the number of clusters formed,
but instead, it ensures that each cluster contains a pre-specified number of units. 
Thus, TC is an effective way of obtaining many clusters, with each containing only a few units.  
Second, TC ensures the formation of a clustering with a small maximum within-group dissimilarity---more precisely, TC finds approximately optimal clustering with respect to a \textit{bottleneck} objective---as opposed to an average or median within-group dissimilarity. 
The bottleneck objective is chosen not only to prevent largely dissimilar units from being grouped together, but also because these types of optimization problems often have approximate solutions that can be found efficiently \citep{hochbaum1986unified}.

More precisely, let $\mathbf B(t^*)$ denote the set of all \textit{threshold clusterings}---those clusterings $\mathbf v$ such that $|V_\ell| \geq t^*$ for each cluster $V_\ell \in \mathbf v$.  
The \textit{bottleneck threshold partitioning problem} (BTPP) is to find the threshold clustering that minimizes the maximum within-cluster dissimilarity.  
That is, BTPP aims to find $\mathbf v^\dagger \in \mathbf B(t^*)$ satisfying:
\begin{equation}
     \max_{V_\ell \in \mathbf v^\dagger} \max_{ij \in V_\ell} d_{ij} = \min_{\mathbf v \in \mathbf B(t^*)}\max_{V_\ell \in \mathbf v} \max_{ij \in V_\ell} d_{ij} \equiv \lambda;
\end{equation}
here, $\lambda$ is the optimal value of the maximum within-cluster dissimilarity.

It can be shown that BTPP is NP-hard, and in fact, no $(2-\epsilon)$--approximation algorithm for BTPP exists unless $\text{P} = \text{NP}$.
However, \citet{higgins2016improving} develop a threshold clustering algorithm (for clarity, the abbreviation TC refers specifically to this algorithm) to find a threshold clustering with maximum within-cluster dissimilarity at most $4\lambda$.
That is, TC is a 4--approximation algorithm for BTPP. 
The time and space requirement for TC are $O(t^*n)$ outside of the construction of a $t^*$--nearest neighbors graph~\citep{higgins2016improving}.
Constructing a nearest neighbor graph is a well-studied problem for which many efficient algorithms  already exist.
At worst, forming a $k$--nearest neighbor graph requires $O(n^2 \log n)$ time~\citep{knuth1998art}; however, if the covariate space is low-dimensional, this construction may only require $O(kn \log n)$ time~\citep{friedman1976algorithm,vaidya1989ano}.
Hence, TC may be used for large datasets, especially when the threshold $t^*$ and the dimensionality of the covariate space are small.

TC uses graph theoretic ideas in its implementation.  
See Appendix~\ref{sec:grthrdef} for graph theory definitions.
TC with respect to a pre-specified minimum cluster size threshold $t^*$ is performed as follows:
\begin{enumerate}
	\item \label{TC_nng} \textbf{(Construct nearest-neighbor subgraph)} Construct a $(t^*-1)$-nearest-neighbors subgraph $NG_{t^*-1}$ with respect to the dissimilarity measure $d_{ij}$ (We use Euclidean distance to measure dissimilarity in this paper).
	\item \label{TC_seed} (\textbf{Choose set of seeds}) Choose a set of units $\mathbf S$ such that
	\begin{enumerate}
         \item For any two distinct units $i, j \in \mathbf S$, there is no walk of length one or two in $NG_{t^*-1}$ from $i$ to $j$.
		\item For any unit $i \notin \mathbf S$, there is a unit $j \in \mathbf S$ such that there exists a walk from $i$ to $j$ of length at most two in $NG_{t^*-1}$.
		\end{enumerate}
	Units in $S$ are known as \textit{seeds}.
	\item \label{TC_group} (\textbf{Grow from seeds})
    For each $\ell \in \mathbf S
    $, form a cluster of units $V^*_\ell$ comprised of unit $\ell$ and all units adjacent to $\ell$ in $NG_{t^*-1}$. 

    \item \label{TC_assign} (\textbf{Assign remaining vertices}) Some units $j$ may not be assigned to a cluster yet. 
    These units are a walk of length two from at least one seed $\ell \in \mathbf S$ in $NG_{t^*-1}$.
    Assign the unassigned units to the cluster associated with seed $\ell$. 
    If there are several choices of seeds, choose the one that with the smallest dissimilarity $d_{\ell j}$. 
\end{enumerate}

The set of clusters $\mathbf v = \{V^*_\ell \}_{\ell \in \mathbf S}$ form a threshold clustering.
Additionally, polynomial-time improvements to this algorithm---for example, in selecting cluster seeds or splitting large clusters---may improve the performance of TC without substantially increasing its runtime.
An implementation of TC can be found in the R package \texttt{scclust}~\citep{scclust}.

\section{Extension of Threshold Clustering}
\label{makereference1.3}
We now describe two extensions of TC: applying TC to instance selection problems and using TC as a preprocessing step for statistical clustering methods.
\subsection{Threshold clustering as instance selection}
\label{makereference1.3.1}
Instance selection methods are used in massive data settings to efficiently scale down the size of the data \citep{leyva2015three}. 
Common techniques for instance selection include subsampling \citep{pooja2013comparative} and constructing \textit{prototypes} \citep{plasencia2014towards}---pseudo data points where each prototype represents a group of similar individual units.  
These methods tend to work quite well for massive data applications. 

\begin{figure}[ht]
    \centering
	\minipage{0.33\textwidth}
	\includegraphics[width=\linewidth,page=6,trim={1.5cm 2.5cm 1cm 2cm},clip]{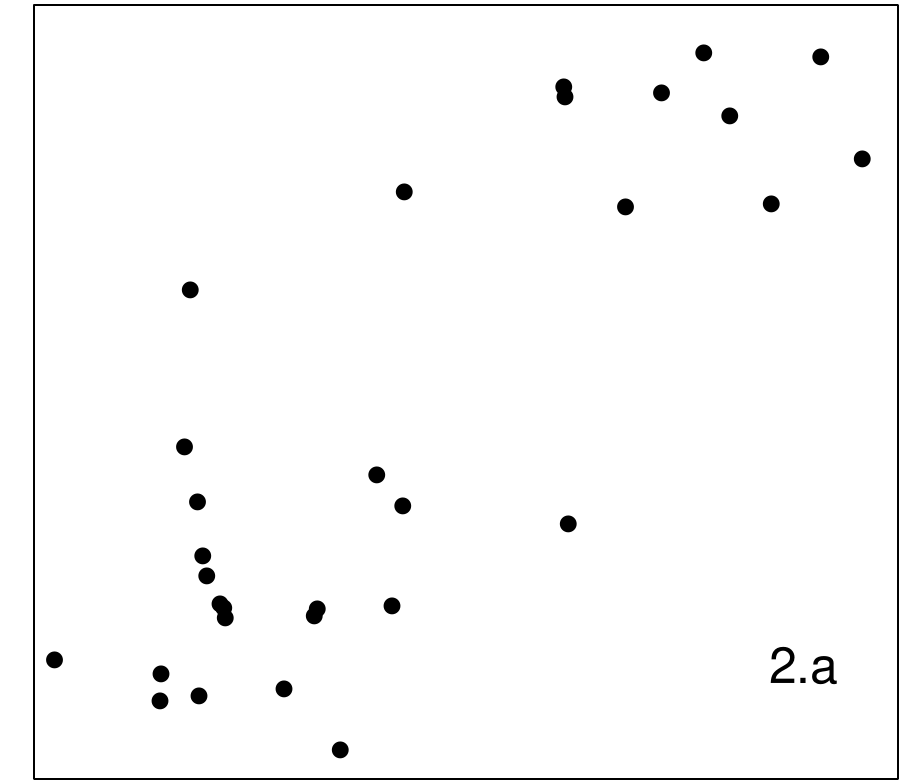}	\label{illu2_1}
	\endminipage
	\minipage{0.33\textwidth}
	\includegraphics[width=\linewidth,page=7,trim={1.5cm 2.5cm 1cm 2cm},clip]{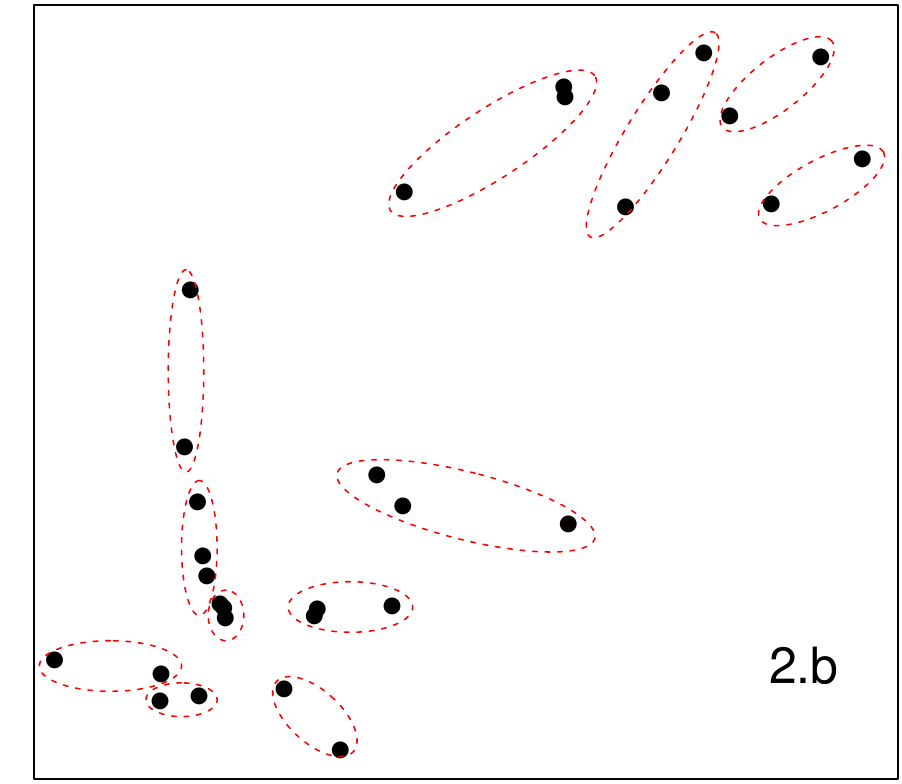}	\label{illu2_2}
	\endminipage
	\minipage{0.33\textwidth}%
	\includegraphics[width=\linewidth,page=8,trim={1.5cm 2.5cm 1cm 2cm},clip]{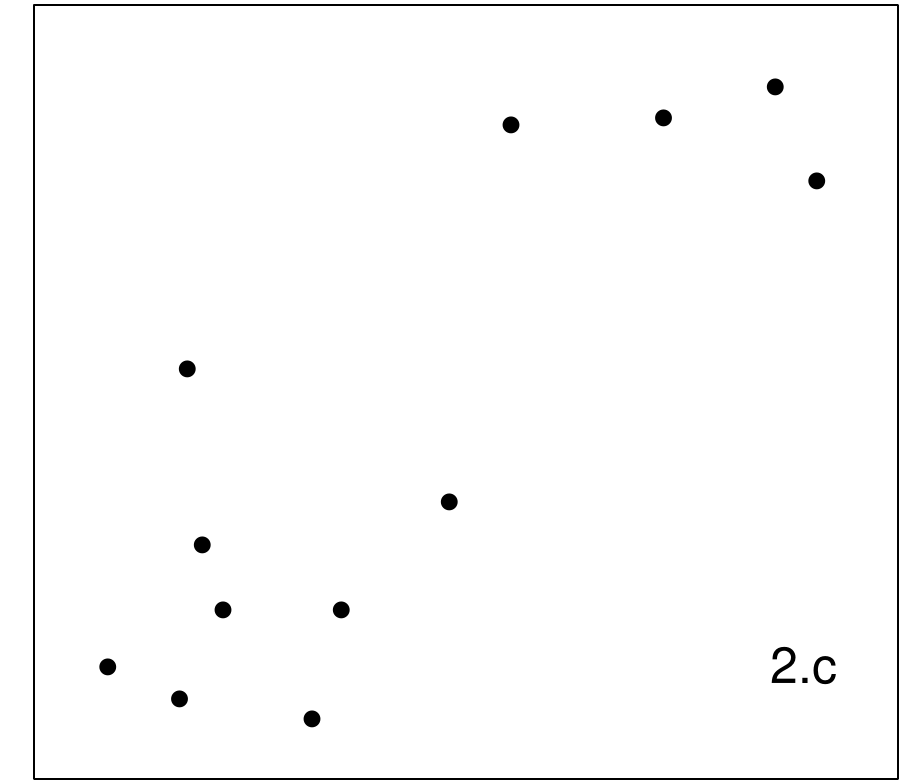}	\label{illu2_3}
	\endminipage\hfill
	\minipage{0.33\textwidth}%
	\includegraphics[width=\linewidth,page=9,trim={1.5cm 2.5cm 1cm 2cm},clip]{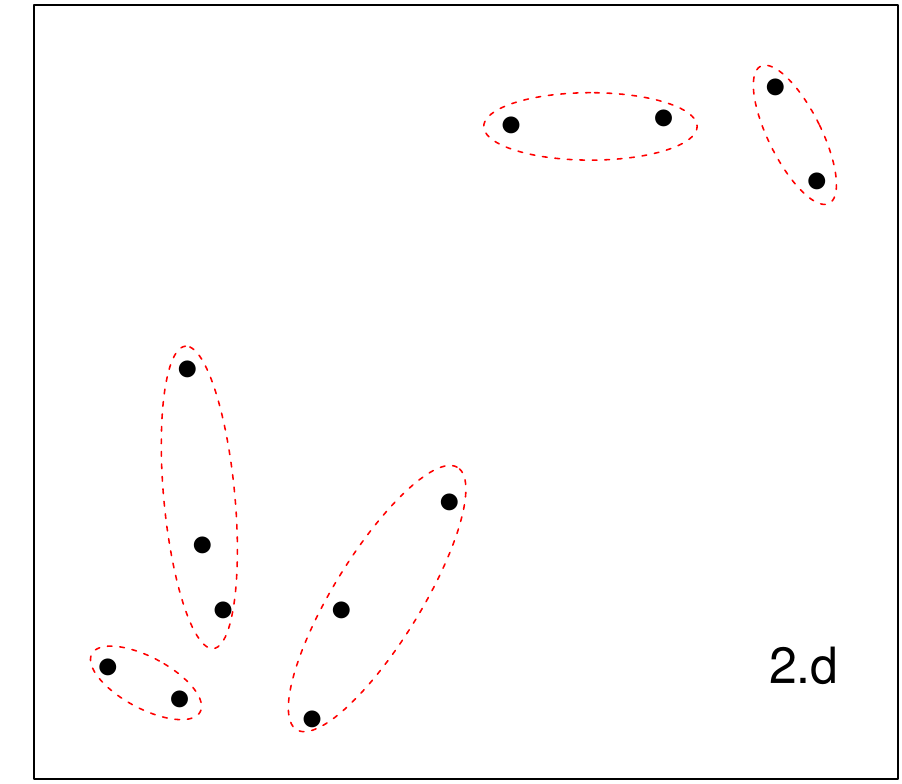}	\label{illu2_4}
	\endminipage
	\minipage{0.33\textwidth}
	\includegraphics[width=\linewidth,page=10,trim={1.5cm 2.5cm 1cm 2cm},clip]{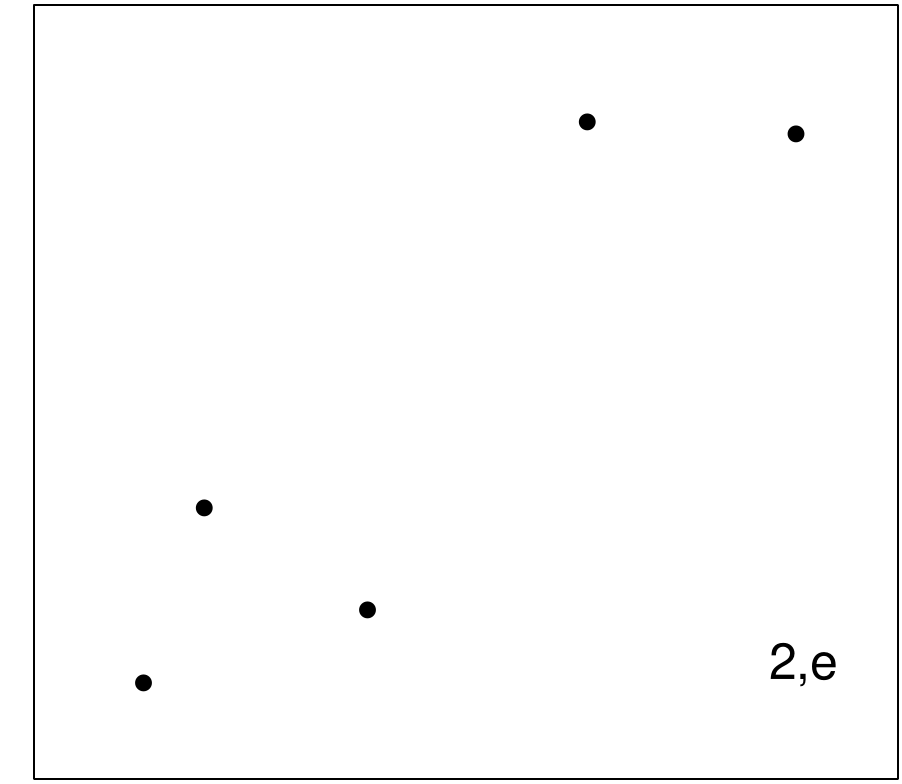}	\label{illu2_5}
	\endminipage
	\caption[An illustration of iterated threshold instance selection (ITIS)]{An illustration of iterated threshold instance selection (ITIS) with threshod $t^* = 2$ on a sample of $n = 30$ data points drawn from a bivariate Gaussian mixture model.  
	In (2.a), each unit is represented by a point. 
	Applying threshold clustering on the data, 12 clusters are formed in (2.b). A prototype is formed for each of the 12 groups in (2.c). In (2.d), apply threshold clustering on the prototype, 5 clusters are formed. 5 prototypes are formed for the 5 clusters, the results are shown in (2.e). For this illustration, We perform two iterations of ITIS.}
	\label{fig_illu_itis}
    \end{figure}

Suppose the researcher desires to reduce the size of the data by a factor of $\alpha$. 
We propose the following method for instance selection, which we call iterated threshold instance selection (ITIS):
	\begin{enumerate}
		\item \label{itis_1} \textbf{(Threshold clustering)} Perform threshold clustering with respect to a small size threshold $t^*$ (for example, $t^* = 2$) on the $n$ data points to form $n^*$ clusters, each containing $t^*$ or more points.
		\item \label{itis_2}\textbf{(Create prototypes)} Compute a center point for each of the $n^*$ groups (for example, centroid or medoid).
		\item \textbf{(Terminate or continue)} 
		\label{itis_3}
		If the data is reduced by a factor of $\alpha$, stop.  
		Otherwise, replace the $n$ data points with the $n^*$ centers and go back to Step~\ref{itis_1}.
	\end{enumerate}
An illustration of the ITIS procedure is given in Figure~\ref{fig_illu_itis}. 


Ultimately, the choice of the number of iterations of ITIS $m$ depends on the researcher.  
For example, a researcher may want to scale down the data as little as necessary in order to perform a computationally intensive statistical procedure on the reduced data.
Additionally, the performance of TC may depend on the dissimilarity measure $d_{ij}$; in our experience, using the standardized Euclidean distance tends to work well.

The running time of $m$ iterations of ITIS is $O(t^*m n \log n)$.
Moreover, since the size of the data is reduced by a factor of $t^*$ with each iteration, the computational bottleneck of ITIS becomes the construction of a $t^*$--nearest-neighbors graph on all $n$ units.
This also suggests that the computation required of ITIS may be drastically improved through the discovery of methods for parallelization of threshold clustering.

The iterative nature of ITIS does have a drawback; with each iteration, the prototype units become less similar to the units comprising the prototype.
In particular, the approximate optimality property of TC may not hold if $m > 1$.
However, simulations suggest that this issue is not severe in massive data settings.
Alternatively, approximate optimality may be preserved by choosing $t^* = \alpha$ and running one iteration of ITIS.  
However, the one iteration version does not seem to be as promising under massive data settings since the runtime of TC increases as $t^*$ increases.
See Appendix~\ref{AppendixA} for details.

\subsection{Iterative Hybridized Threshold Clustering}
\label{makereference1.3.2}
\begin{figure}[ht]
    \centering
	\minipage{0.33\textwidth}
	\includegraphics[width=\linewidth,page=13,trim={1.5cm 2.5cm 1cm 2cm},clip]{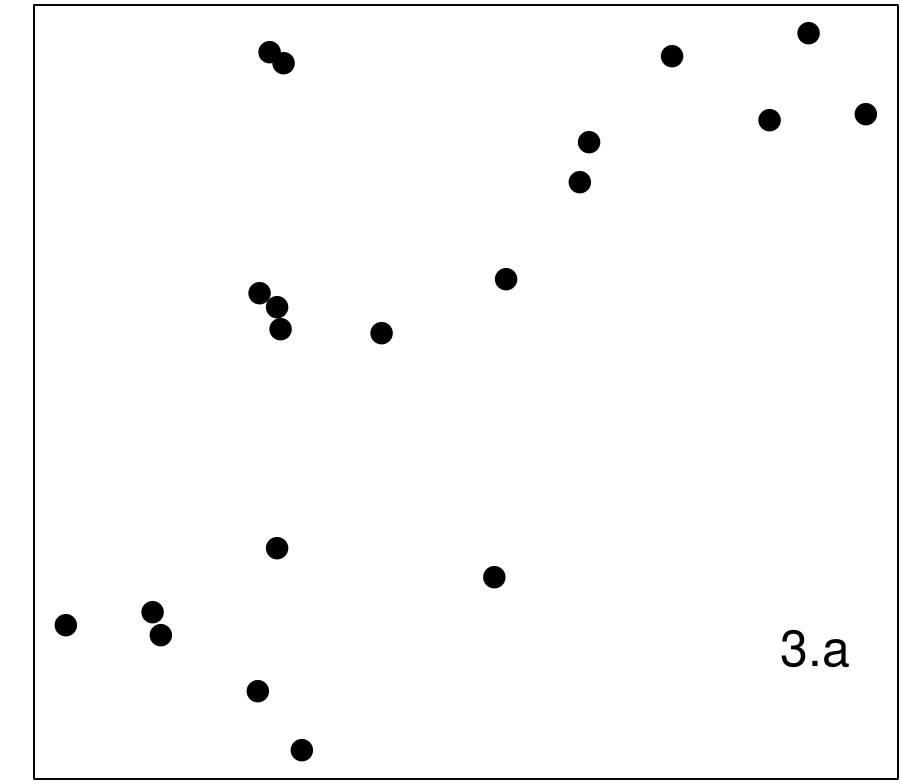}	\label{illu3_1}
	\endminipage
	\minipage{0.33\textwidth}
	\includegraphics[width=\linewidth,page=14,trim={1.5cm 2.5cm 1cm 2cm},clip]{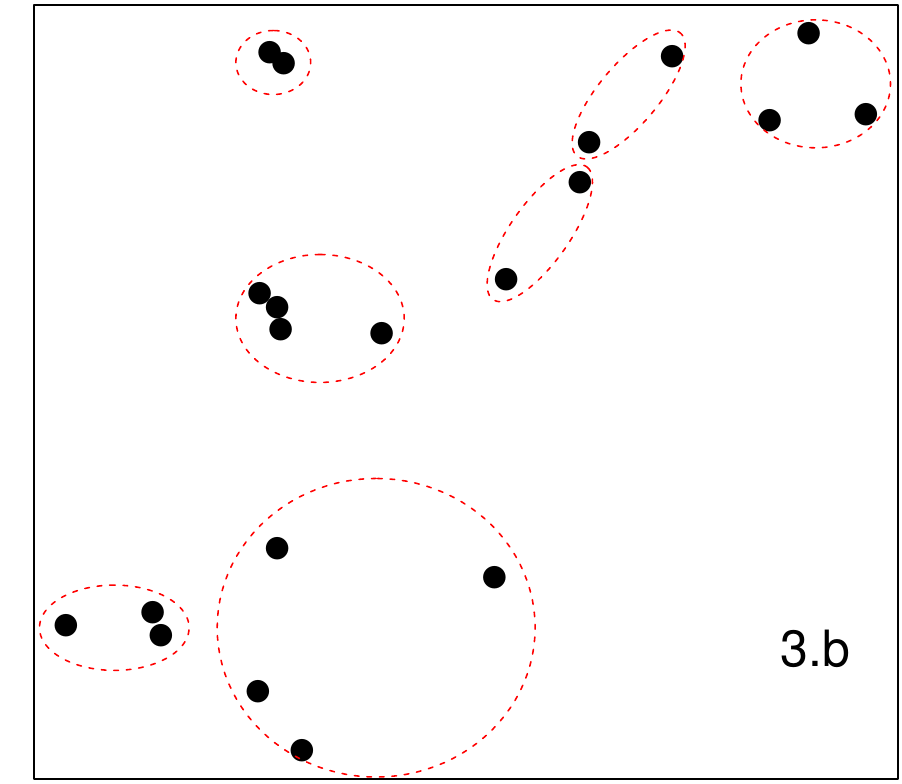}	\label{illu3_2}
	\endminipage
	\minipage{0.33\textwidth}%
	\includegraphics[width=\linewidth,page=15,trim={1.5cm 2.5cm 1cm 2cm},clip]{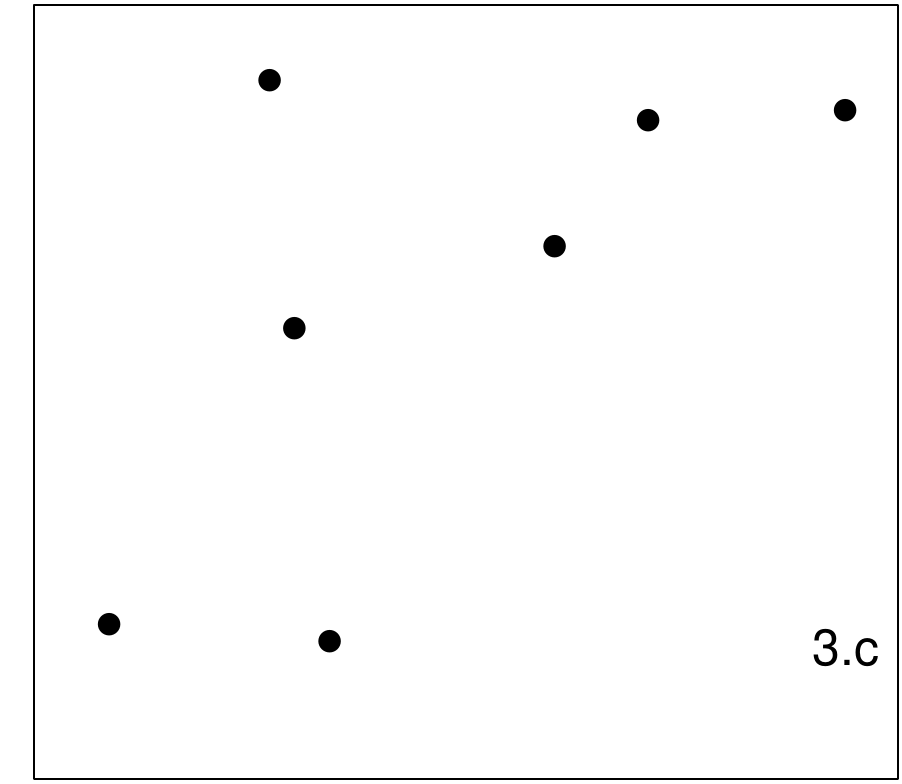}	\label{illu3_3}
	\endminipage\hfill
	\minipage{0.33\textwidth}%
	\includegraphics[width=\linewidth,page=16,trim={1.5cm 2.5cm 1cm 2cm},clip]{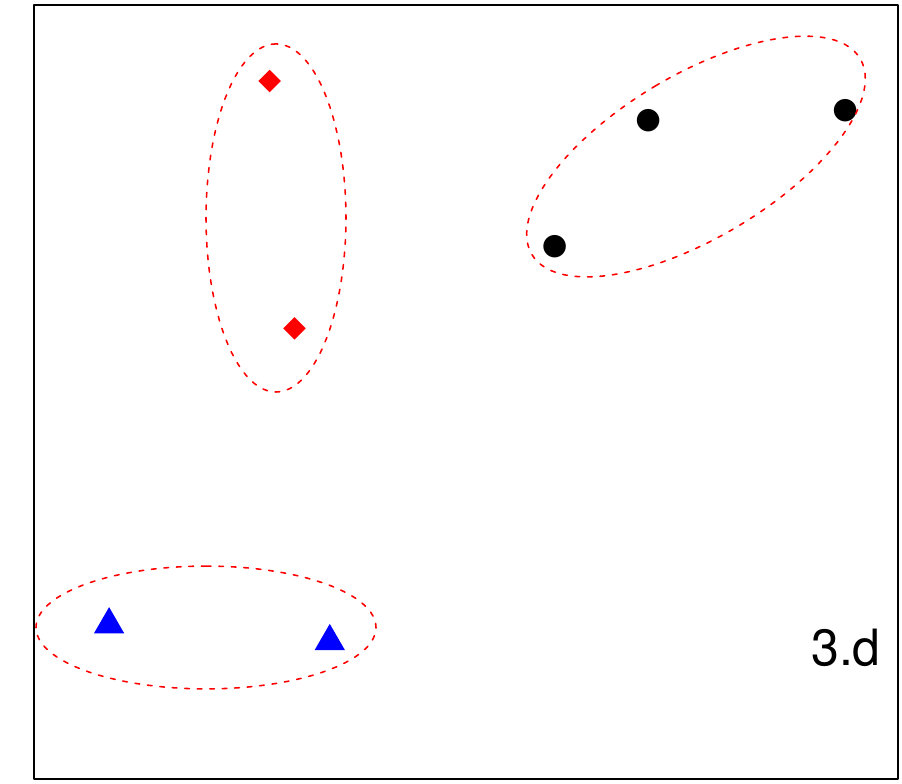}	\label{illu3_4}
	\endminipage
	\minipage{0.33\textwidth}
	\includegraphics[width=\linewidth,page=17,trim={1.5cm 2.5cm 1cm 2cm},clip]{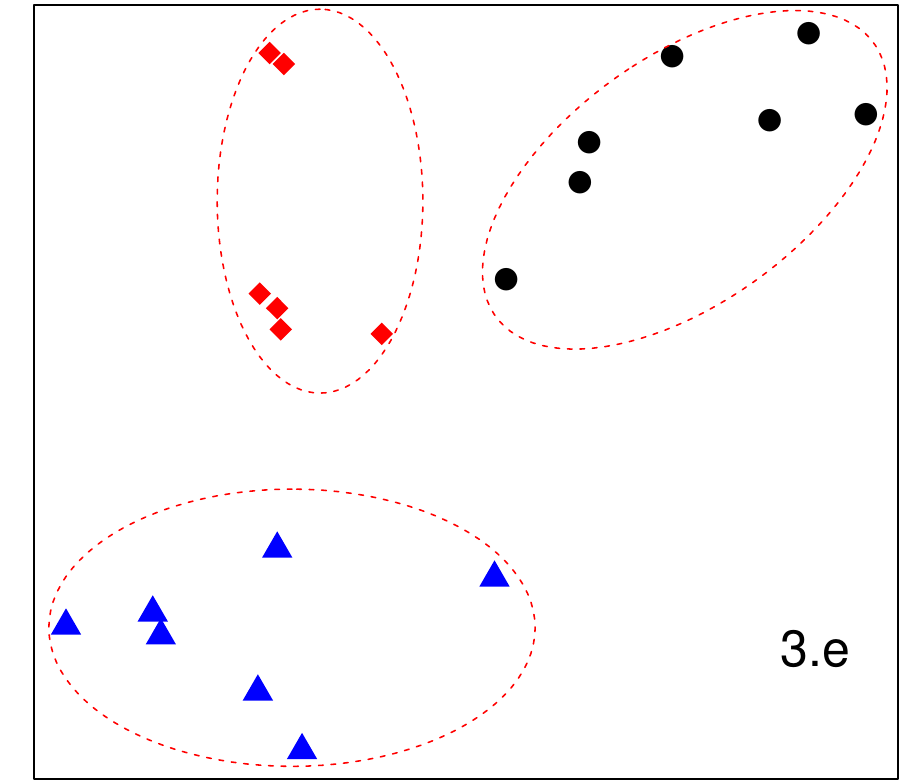}	\label{illu3_5}
	\endminipage
	\caption[An illustration of iterative hybridized threshold clustering with $k$-means]{An illustration of hybridized threshold clustering with $k$-means on bivariate data: $n=20, k=3,t^*=2$. 
	In graph (3.a), each unit is represented by a point. Applying threshold clustering on the data, 7 clusters are formed in (3.b). A prototype is formed for each of the 7 groups in (3.c). In (3.d), $k$-means clustering is performed on the prototypes. The "backed out" final clustering results are given in (3.e).}
	\label{fig_illu_ihtc}
    \end{figure}

Often, researchers would like to use certain clustering methods (for example, $k$-means, HAC, etc.) because of favorable or familiar properties of these clustering methods.
However, under massive data settings, using such clustering techniques may not be feasible because of prohibitive computational costs.
We propose the following method for using ITIS as a pre-processing step on all $n$ units to allow for the use of more sophisticated clustering methods.
We call this procedure Iterative Hybridized Threshold Clustering (IHTC).
It works as follows:
\begin{enumerate}
	\item \label{ihtc_1}\textbf{(Create prototypes)} Perform iterative threshold instance selection with respect to a size threshold $t^*$ on the $n$ data points $m$ times to form prototypes.
	\item \textbf{(Cluster prototypes)}\label{ihtc_2}  Cluster the prototypes (for example, using $k$-means) obtained by~\ref{ihtc_1}.
    \item \textbf{("Back out" assignment)} \label{ihtc_3} For each prototype, determine which of the original $n$ units contributed to the formation of that prototype and assign these units to the cluster belonging to the prototype.
\end{enumerate}
Figure~\ref{fig_illu_ihtc} gives an illustration of IHTC with $k$-means.

IHTC reduces the size of the data by a factor of $(t^*)^m$, which can improve the efficiency of the original clustering algorithm. 
Additionally, IHTC also prevents overfitting of individual points, which may lead to a more effective clustering regardless with just a couple iterations of IHTC. Specifically, for $m$ iterations of ITIS at size threshold $t^*$, IHTC ensures that each cluster contains at least $(t^*)^m$ units. 
Finally, we note that IHTC may be applied to most other clustering algorithms---not just $k$-means or HAC.

\section{Simulation}
\label{makereference1.4}
We first demonstrate properties of IHTC using simulated data.   
We apply IHTC to samples of varying sizes where each data point is sampled from a mixture of three bivariate Gaussian distributions.  
Specifically, samples are drawn independently from a distribution with pdf:
\begin{equation*}
f(\mathbf{x})=0.5 p(\mathbf{x}|\mathbf{\mu_1, \Sigma_1})+0.3 p(\mathbf{x}|\mathbf{\mu_2, \Sigma_2})+0.2 p(\mathbf{x}|\mathbf{\mu_3, \Sigma_3})
\end{equation*}
where $p(\mathbf{x}|\mathbf{\mu_j, \Sigma_j})$ is the pdf of a Gaussian with parameters $\mathbf{\mu_j}$ and $\mathbf{\Sigma_j}$, $j=1,2,3$, 
\begin{equation*}
\mathbf{\mu_1}=\begin{bmatrix} 1 \\ 2 \end{bmatrix}, 
\mathbf{\mu_2}=\begin{bmatrix} 7 \\ 8 \end{bmatrix}, 
\mathbf{\mu_3}=\begin{bmatrix} 3 \\ 5 \end{bmatrix},
\end{equation*}
\begin{equation*}
\mathbf{\Sigma_1}=\begin{bmatrix} 1 & 0 \\ 0 & 0.5 \end{bmatrix}, 
\mathbf{\Sigma_2}=\begin{bmatrix} 2 & 0 \\ 0 & 1 \end{bmatrix}, 
\mathbf{\Sigma_3}=\begin{bmatrix} 3 & 0 \\ 0 & 4 \end{bmatrix}. 
\end{equation*}

We use Euclidean distance as our dissimilarity measure.  
The data size $n$ varies between $10^4$ and $10^8$ observations and each setting is replicated 1000 times.
For each simulation, we record the run time in seconds and memory usage in megabytes for the whole procedure.

Algorithms are implemented in the R programming language.
We use the \texttt{scclust} package to perform threshold clustering~\citep{scclust}. 
The default $\texttt{kmeans}$ and $\texttt{hclust}$ functions in R are used for $k$-means and hierarchical agglomerative clustering respectively. 
This simulation was performed on the Beocat Research Cluster at Kansas State University, which is funded in part by NSF grants CNS-1006860, EPS-1006860, and EPS-0919443~\citep{BEOCAT}.
We perform simulations on a single core Intel Xeon E5-2680 v3 with 2.5 GHz processor with 30 GM of RAM at maximum.

\subsection{IHTC with K-means Clustering \label{makereference1.4.1}}
\begin{figure}[H]
\centering
	\minipage{0.5\textwidth}	\includegraphics[width=\linewidth,page=1]{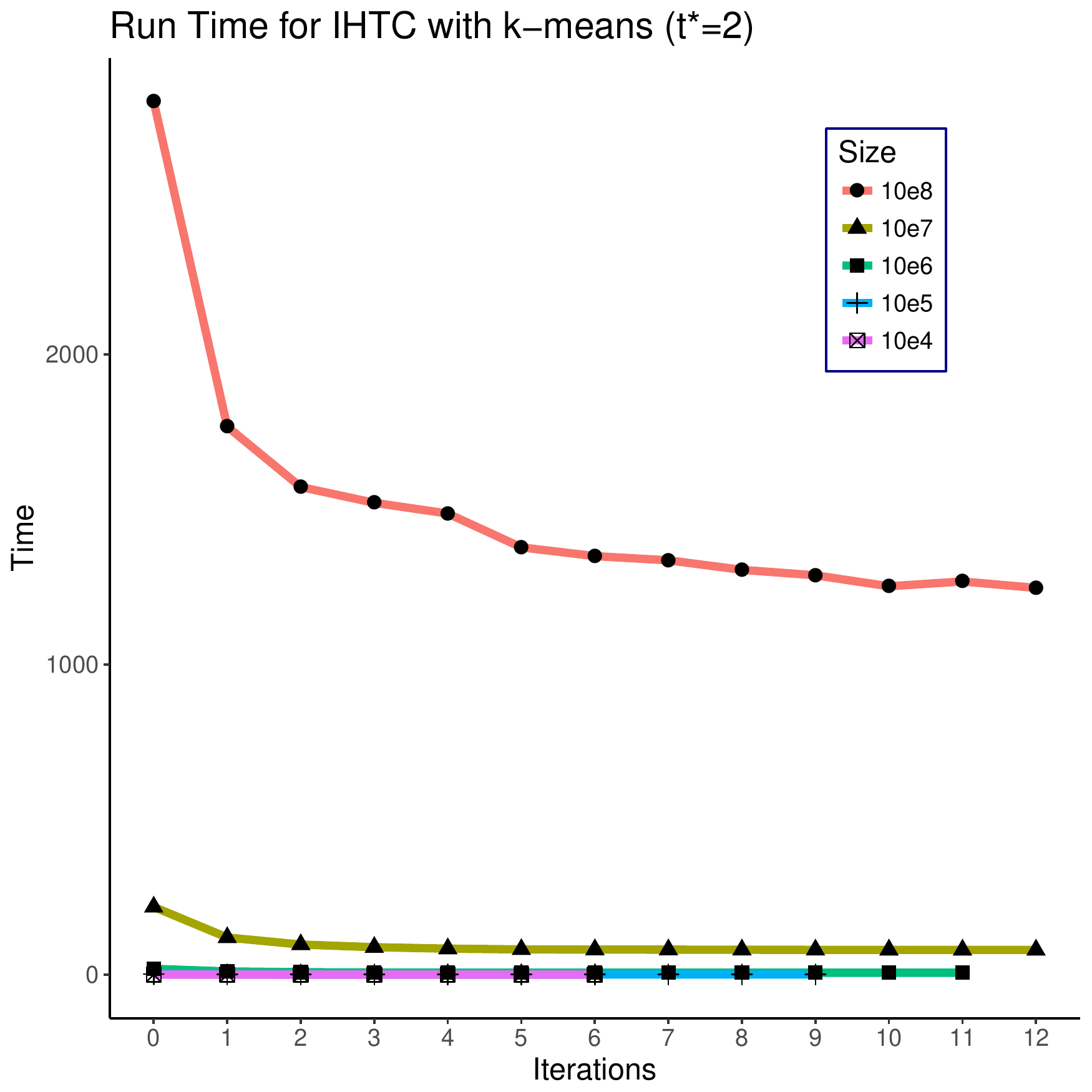}
	\label{fmul2_km_time}
	\endminipage
	\minipage{0.5\textwidth}
\includegraphics[width=\linewidth,page=2]{simulation/fmul2_km_avg_merge.pdf}
	\label{fmul2_km_logtime}
	\endminipage\hfill
	\minipage{0.5\textwidth}	\includegraphics[width=\linewidth,page=3]{simulation/fmul2_km_avg_merge.pdf}
	\label{fmul2_km_mem}
	\endminipage
	\minipage{0.5\textwidth}
\includegraphics[width=\linewidth,page=4]{simulation/fmul2_km_avg_merge.pdf}
	\label{fmul2_km_logmem}
	\endminipage
	\caption{Run time and memory usage of IHTC with $k$-means clustering ($k=3,t^*=2$).}
	\label{fig_ihtc_km}
\end{figure}

\begin{figure}[ht]
\centering
	\includegraphics[width=0.5\linewidth,page=5,  trim={0.1cm 0.6cm 0.6cm 0.75cm},clip]{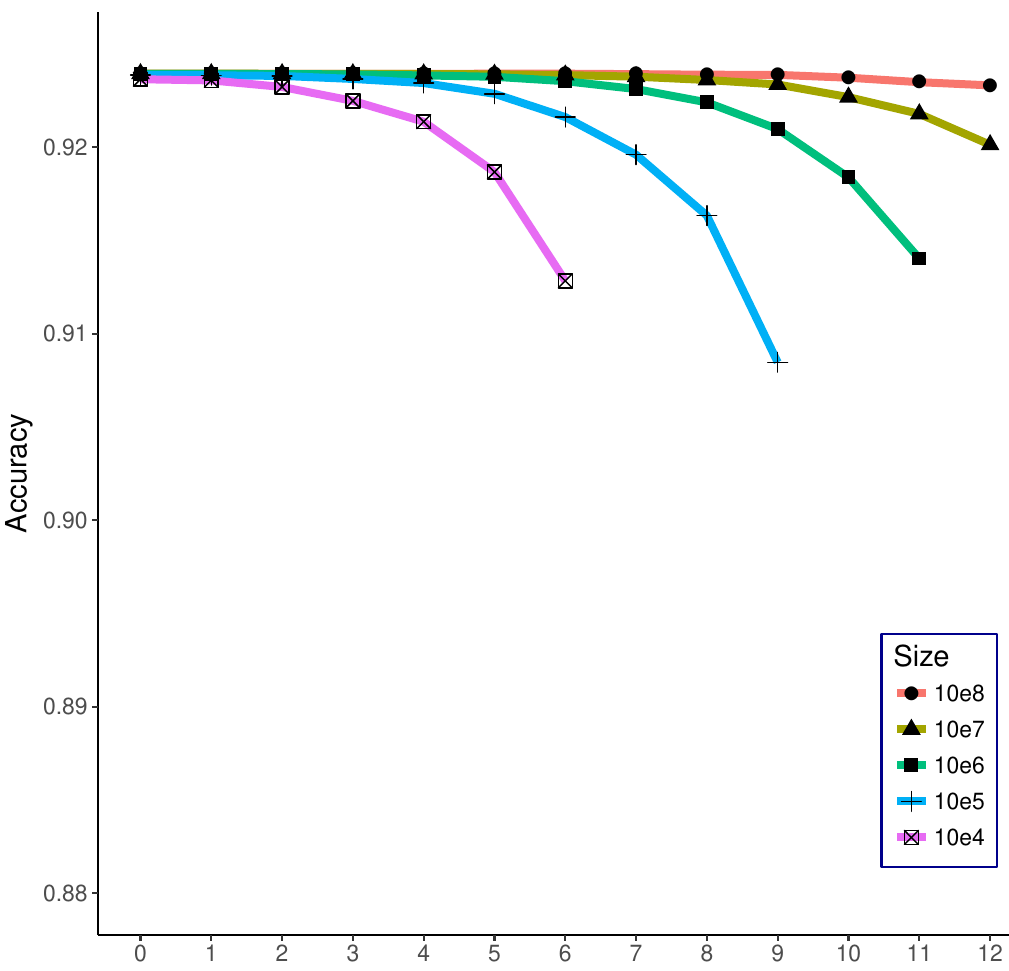}
	\caption{Prediction accuracy of IHTC with $k$-means clustering ($k=3,t^*=2$).}
	\label{fig_ihtc_accu}
\end{figure}

We perform IHTC with $k$-means with threshold $t^*=2$ and number of clusters $k=3$.
The run time and memory usage are in Figure~\ref{fig_ihtc_km} and Table~\ref{t_fmul2_km}. 
Because we simulated data from a Gaussian mixture model, each cluster roughly should correspond to a different Gaussian distribution.
Hence, we can use prediction accuracy to measure the performance of our methods. 
The prediction accuracy is the number of units correct clustered divided by data size $n$. 
The prediction accuracy for IHTC with $k$-means are in Figure~\ref{fig_ihtc_accu} and Table~\ref{t_fmul2_km}.
The first point of each curve (iteration $m=0$) indicates the performance without pre-processing of the data; that is $m=0$ indicates where only the original clustering algorithm is applied to the data.  

From Figures~\ref{fig_ihtc_km} and~\ref{fig_ihtc_accu}, and from Table~\ref{t_fmul2_km}, we find that using IHTC with $k$-means decreases the runtime and memory required compared to without IHTC. 
After one iteration, the runtime and memory usage decreases by about half while the prediction accuracy remains the same. 
As the number of iterations increases, the additional improvements in runtime and memory usage decrease. 
After several iterations, runtime and memory usage tend to level off and prediction accuracy slowly decreases. 

\begin{landscape}
\begin{table}
	\centering
	\resizebox{\linewidth }{!}{%
	\begin{tabular}{ c | c c c c c | c c c c c | c c c c c }
		Iteration & \multicolumn{5}{c}{Run Time (second)} & \multicolumn{5}{c}{Memory (Mb)} & \multicolumn{5}{c}{Accuracy}\\
		 ($m$) & $10^4$ & $10^5$ & $10^6$ & $10^7$ & $10^8$  & $10^4$ & $10^5$ & $10^6$ & $10^7$ & $10^8$ &$10^4$ & $10^5$ & $10^6$ & $10^7$ & $10^8$\\
		\hline
		0 &  0.143 & 1.613 & 18.773 & 218.43 & 2815 & 19.39 & 241.64 & 2556 & 27540 & 279346 & 0.9236 & 0.9239  & 0.9239  & 0.9239  & 0.9239\\
		1 &  0.084 & 0.909 & 10.337 & 119.86 & 1767 &  8.70 &  99.14 & 1019 & 11097 & 110467 & 0.9236 & 0.9239  & 0.9239  & 0.9239  & 0.9239\\
		2 &  0.072 & 0.647 &  7.886 &  97.07&  1572 &   3.99 &  44.00 &  488 &  5462 &  54773  & 0.9232 &  0.9238 &  0.9239  & 0.9239  & 0.9239\\
		3 &  0.058 & 0.550 &  6.975 &  88.52&  1522 &   1.76 &  23.55 &  253 &  3104 &  31764 &  0.9225  & 0.9237  & 0.9239  & 0.9239  & 0.9239\\
		4 &  0.053 & 0.502 &  6.534 &  83.86&  1487 &   0.97 &  14.19 &  166 &  2150 &  21962 &  0.9214  & 0.9234  & 0.9238  & 0.9239 &  0.9239\\
		5 &  0.053 & 0.497 &  6.332 &  81.46&  1378 &   0.81 &   8.62 &  130 &  1761 &  17757 &  0.9187  & 0.9229  & 0.9238  & 0.9239  & 0.9239\\
		6 &  0.051 & 0.487 &  6.272 &  80.90&  1350 &   0.81  &  7.94 &  112 &  1614 &  16478 &  0.9128  & 0.9216  & 0.9235 &  0.9239  & 0.9239\\
		7 &  - & 0.487 &  6.263  & 80.62 & 1336 &   - &   7.69 &  106  & 1560 &  15813 &  -  & 0.9196  & 0.9231  & 0.9238 &  0.9239\\
		8 &  - & 0.490 &  6.254  & 80.28&  1305 &   -  &  7.56 &  105 &  1561 &  16005 &  -  & 0.9163  & 0.9224  & 0.9236  & 0.9239\\
		9  &      - & 0.490 &  6.243  & 80 & 1288 &       - &   7.91 &  105 &  1574 &  16099 &    -  & 0.9085  & 0.9210  & 0.9234  & 0.9239\\
		10  &     - & - &  6.245  & 79.95&  1252 &     - &   - &  106 &  1596 &  16512 &    -  & -  & 0.9184  & 0.9227  & 0.9237\\
		11  &     -   &    - &  6.246 &  79.75 & 1268 &     - &       - &  109 &  1630 &  16587 &    -   &    -  & 0.9140  & 0.9218  & 0.9235\\
		12  &     -   &    -   &     - &  79.72 & 1247 &     -  &      -  &      - &  1662 &  17036 &    -   &    -   &    -  & 0.9201  & 0.9233\\
		\hline
	\end{tabular}
	}
	\caption{Cluster performance of IHTC with $K$-means ($k=3,t^*=2$).}
	\label{t_fmul2_km}
\end{table}
\end{landscape}

\subsection{IHTC with HAC}
\label{makerference 1.4.2}
\begin{figure}[H]
\centering
	\minipage{0.5\textwidth}	\includegraphics[width=\linewidth,page=1]{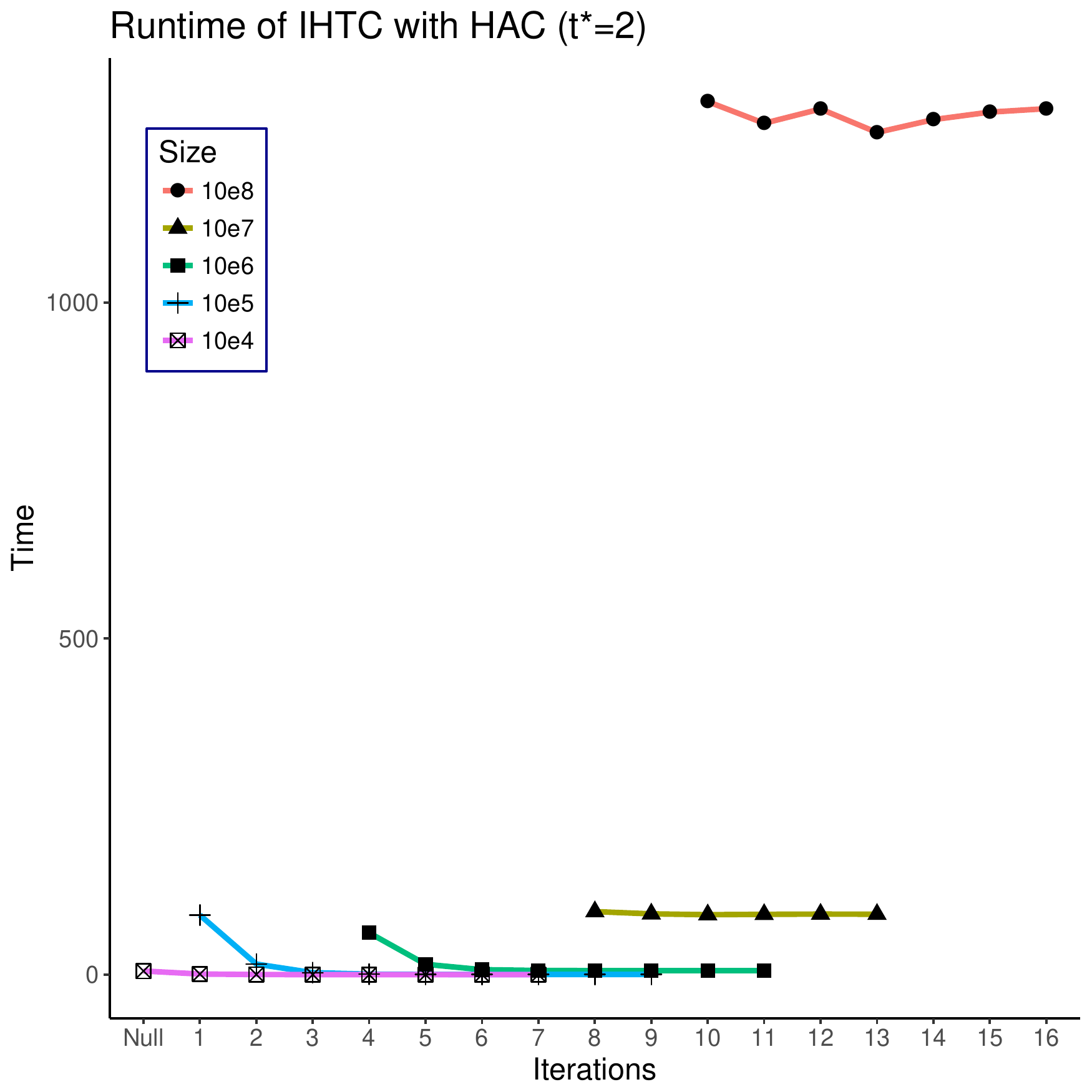}
	\label{fmul2hac_time}
	\endminipage
	\minipage{0.5\textwidth}
\includegraphics[width=\linewidth,page=2]{simulation/fmul2_hac_merge.pdf}
	\label{fmul2hac_logtime}
	\endminipage\hfill
	\minipage{0.5\textwidth}	\includegraphics[width=\linewidth,page=3]{simulation/fmul2_hac_merge.pdf}
	\label{fmul2hac_mem}
	\endminipage
	\minipage{0.5\textwidth}
\includegraphics[width=\linewidth,page=4]{simulation/fmul2_hac_merge.pdf}
	\label{fmul2hac_logmem}
	\endminipage
	\caption{Run time and memory usage of IHTC with HAC ($t^*=2$).}
	\label{fig_ihtc_hac}
\end{figure}

\begin{figure}[H]
\centering
	\includegraphics[width=0.5\linewidth,page=5]{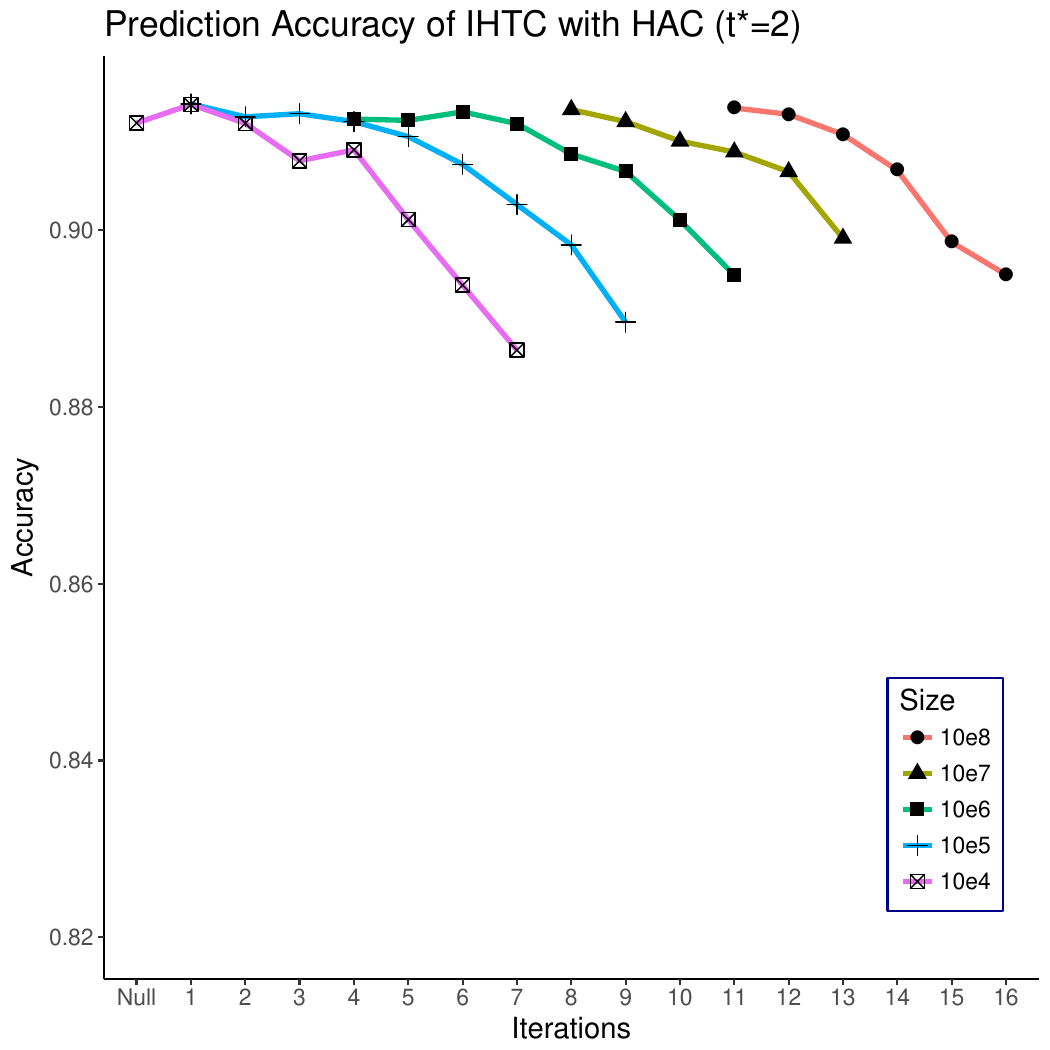}
	\caption{Prediction accuracy of IHTC with HAC ($k=3,t^*=2$).}
	\label{fig_ihtc_hacaccu}
\end{figure}

HAC is an computationally expensive algorithm.
For example, if data size exceed 65,536 datapoints, the $\texttt{hclust}$ function in R will throw an error. 
We applied IHTC with HAC and consider with respect to threshold $t^*=2$. 
Comparing the performance with pre-processing and without pre-processing, we found the reduction for runtime and memory requirement is dramatic when we apply IHTC with HAC, while prediction accuracy falls slightly. 
As the number iterations increase, the improvements in runtime and memory usage decrease, and after several iterations, runtime and memory tend to a certain stable value and prediction accuracy decreases. 
Thus, the number of iterations to perform is an unsolved problem. 
Figures~\ref{fig_ihtc_hac} and~\ref{fig_ihtc_hacaccu}, and
Table~\ref{t_fmul2_hac} give the results of IHTC with HAC.

\begin{table}[ht]
	\centering
	\resizebox{\textwidth}{!}{%
	\begin{tabular}{ c | c c c c c | c c c c c | c c c c c}
		Iterations& \multicolumn{5}{c}{Run Time (second)} & \multicolumn{5}{c}{Memory (Mb)} & \multicolumn{5}{c}{Accuracy}\\
		$m$ & $10^4$ & $10^5$ & $10^6$ & $10^7$ & $10^8$ & $10^4$ & $10^5$ & $10^6$ & $10^7$ & $10^8$ & $10^4$ & $10^5$ & $10^6$ & $10^7$ & $10^8$\\
		\hline
		Null  &  5.425   &   -  &  -   &   -   &  - &  796.46   &    -   &  -   &  -   &   - &  0.9122   &    -   &    -    &   -  &    -\\
		1  &  0.940  & 88.619   &  -   & -  &  - &  177.27  & 20193.0   & -  &   -  &   - &  0.9142 & 0.9143    &   -    &   -   &    -\\
		2  &  0.220 & 15.755   & -   & -  & -  &   20.49  &  3537.4    &  -  &   -  &    - &  0.9121 &  0.9129   &    -   &    -   &    -\\
		3  &  0.073  &  2.992    &-   &  - & -  &    7.01  &   459.9    &  -  & -   &   -&  0.9079 & 0.9132    &   -    &   -   &   -\\
		4  &  0.048  &  0.752  & 62.28  & -  & -  &  1.11  &   117.9  & 10891  & -  &   -&  0.9091  & 0.9123  & 0.9126   &    -  &     -\\
		5  &  0.044  &  0.393  & 15.41  & -  & - &  0.73   &   28.7  &  1940  &  -  &   - &  0.9012 & 0.9106  & 0.9124    &   -    &   -\\
		6  &  0.045  &  0.350  &  7.44  &  -  &  - &  0.74  &  11.2 &  435  &  - &   -&  0.8938  & 0.9075  & 0.9134    &   -  &     -\\
		7  &  0.044  &  0.343  &  6.21  &  -  &  - &  0.74    & 9.3  &  165  &  -   &  - &  0.8864  & 0.9029  & 0.9121   &    -   &    -\\
		8 &   -  &  0.342  &  5.99  & 93.8  &  - & -&  9.3  & 120  & 2405   &   -&  - & 0.8984  & 0.9086  & 0.9137   &    -\\
		9  &  - & 0.342  &  5.98  & 90.4 &  - &   - &  9.4  & 115  & 1648   &   -& -  & 0.8896  & 0.9067 & 0.9123    &   -\\
		10   &   -  &  -  &  5.99  & 89.2 &  1299 &  -  &   -  &  118  & 1564 & 19113  &    - & -  & 0.9012  & 0.9101 &  0.9159\\
		11   &  -  &   -  &  6.04  & 89.7 & 1267 &  -    &  - & 123  & 1523  & 16932&    -   &    -  & 0.8949  & 0.9089  & 0.9139\\
		12   &  -  &  -   &   -  & 90.1 & 1288 &   -    &   -    &    -  & 1579  & 16760 &    -   &    -   &    -  & 0.9066 & 0.9131\\
		13   &  -  &    -   &    - & 89.9  & 1253 &  -   &   -    &    -  & 1604  & 17041&    -   &    -   &    -  & 0.8991  & 0.9109\\
		14   &   -  &  -   &   -   &   -  & 1272 &  -     &  -    &    -  &  -  & 17505 &    -   &    -   &    -   &    -  & 0.9068\\
		15   &   -   & - &  -  &   -  & 1283 &    -   &     -     &   -  &   -  & 17784 &    -   &    -   &   -    &   -  & 0.8987\\
		16    &  -   & - &  -   &  -  & 1288 &  -      &    -     &   -   &  -  & 18220&    -   &   -     &  -   &    -  & 0.8950\\
		\hline
	\end{tabular}}
	\caption{Cluster performance for IHTC with HAC ($t^*=2$)}
	\label{t_fmul2_hac}
\end{table}

\section{Experiments}
\label{makereference1.5}
In this section, we will use our threshold clustering algorithm performance on several publicly available datasets. 
A brief description of each dataset can be found in Table~\ref{data_detail}. 
The Euclidean distance is used to measure the dissimilarity and principal component analysis \citep{hotelling1933analysis, friedman2001elements} is used for each dataset to perform feature selection. 
The number of classes ($k$) is chosen by the elbow of plot of within-cluster sum of squared distances for different $k$. 
All experiments run on a single CPU core laptop.

\begin{table}[ht]
\centering
\begin{tabular}{| c | c | c | c |}
\hline
    Name & Instances & Attributes & Classes  \\
    \hline\hline
    PM 2.5 \citep{pm25} & 41757 & 5 & 4 \\
    Credit Score \citep{score} & 120269 & 6 & 5 \\
    Black Friday \citep{blackfriday} & 166986 & 7 & 4 \\
    Covertype \citep{Blackard99comparativeaccuracies} & 581012 & 6 & 7\\
    House Price \citep{zillow} & 2885485 & 5 & 5\\
    Stock \citep{brogaard2014high, carrion2013very} & 7026593 & 5 & 7 \\
    \hline
\end{tabular}
\caption{Data description.}
\label{data_detail}
\end{table}

\begin{figure}[H]
\centering
\includegraphics[width=\linewidth,page=2]{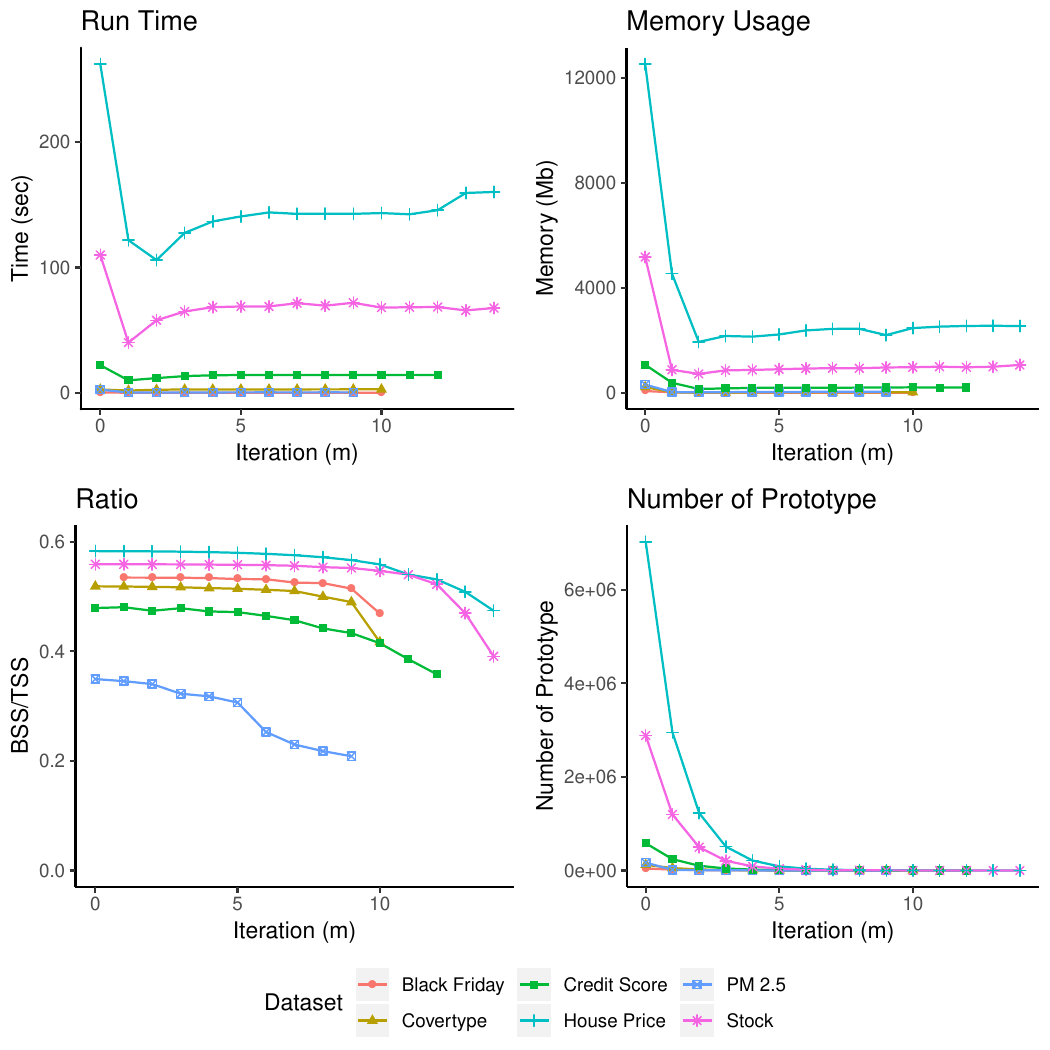}
	\caption{Cluster performance for IHTC with K-means.}
	\label{fig_data1}
\end{figure}

\begin{figure}[H]
\centering
\includegraphics[width=\linewidth,page=2]{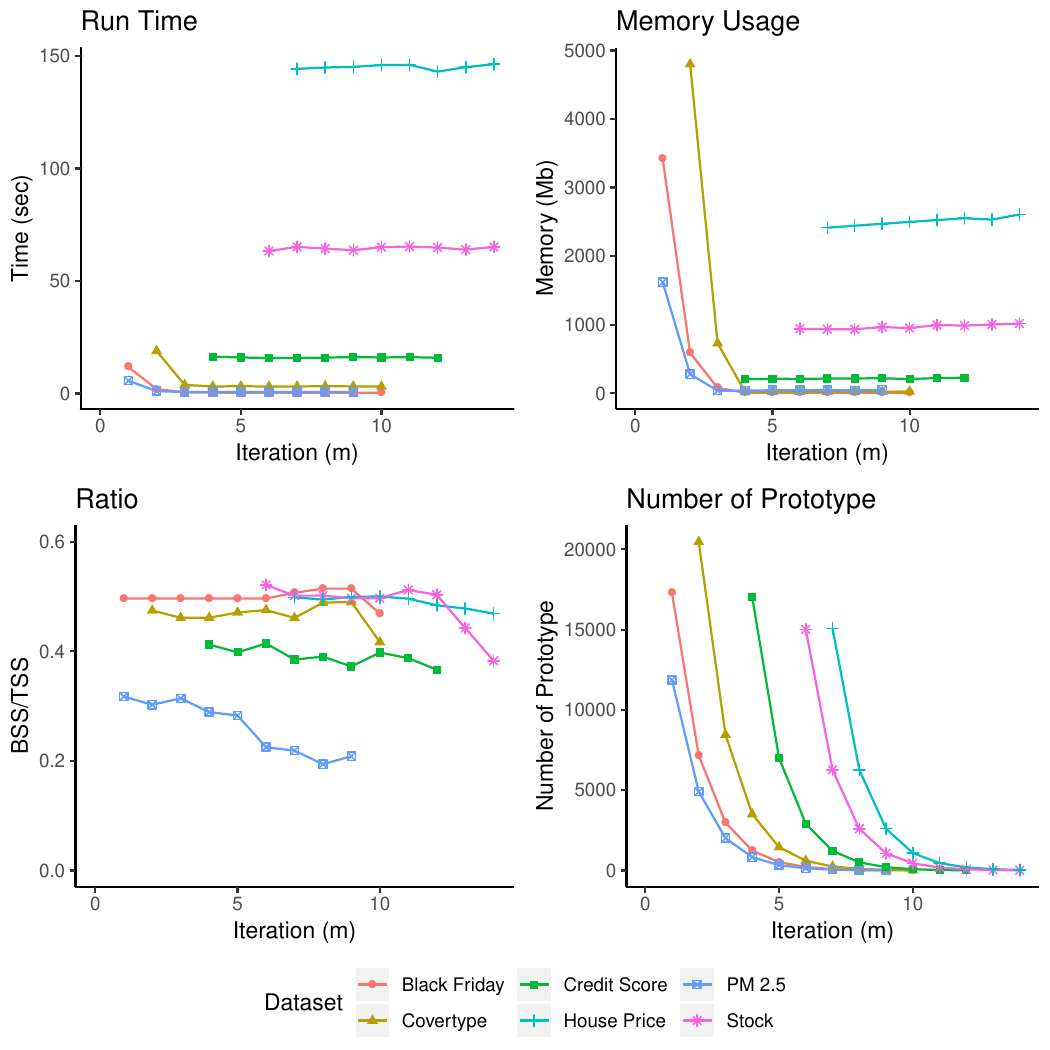}
	\caption{Cluster performance for IHTC with HAC.}
	\label{fig_data2}
\end{figure}

Tables~\ref{data_fkm},~\ref{data_fhac1}, and~\ref{data_fhac2}, and Figures~\ref{fig_data1} and~\ref{fig_data2} demonstrate results of IHTC with $k$-means and HAC for these datasets. 
Iteration $m=0$ indicates the performance when no pre-processing was conducted; only $k$-means clustering was applied to the data. 
BSS/TSS is the ratio of between cluster sum of squares and total sum of squares. 
Larger ratio value indicates better cluster performance. 
The number of prototypes is the size of the reduced data after $m$ iterations. 
We found that using IHTC with $k$-means or HAC decreases the runtime and memory required compared to without IHTC. 
After one iteration, the runtime and memory usage decreases by about half while preserving the value of the BSS/TSS ratio. 
As the number of iterations becomes large, the improvements in runtime and memory usage decrease and the BSS/TSS ratio decreases slowly. 

To demonstrate the versitility of IHTC, we also consider its performance with the clustering method DBSCAN \citep{ester1996density}. 
The results of DBSCAN are contained in Appendix~\ref{AppendixB}.

\begin{landscape}
\begin{table}[H]
	\centering
	\resizebox{\linewidth }{!}{%
	\begin{tabular}{ c | c c c c | c c c c | c c c c | c c c c}
		 Name& \multicolumn{4}{c}{Run Time (second)} & \multicolumn{4}{c}{Memory Usage (Mb)} &\multicolumn{4}{c}{BSS/TSS} & \multicolumn{4}{c}{Number of Prototypes}\\
		  & $m=0$ & $m=1$ & $m=2$ & $m=3$ & $m=0$ & $m=1$ & $m=2$ & $m=3$ & $m=0$ & $m=1$ & $m=2$ & $m=3$  & $m=0$ & $m=1$ & $m=2$ & $m=3$\\
		\hline
		PM 2.5 & 0.636 & 0.282 & 0.232 & 0.268 & 71.28 & 25.69 & 3.7 & 5.91 & 0.5347 & 0.5346 & 0.5345 & 0.5344 & 41757 & 17281 & 7166 & 2984\\
		Credit Score & 2.902 & 2.046 & 2.696 & 2.904 & 224.55 & 23.9 & 13.95 & 17.65 & 0.5187 & 0.5184 & 0.5178 & 0.5169 & 120269 & 49669 & 20471 & 8456\\
		Black Fridy & 2.802 & 0.468 & 0.522 & 0.554 & 317.05 & 32.95 & 24.5 & 28.91 & 0.3493 & 0.3456 & 0.3402 & 0.3226 & 166986 & 11868 & 4914 & 2017\\
		Covertype & 22.244 & 10.184 & 11.968 & 13.562 & 1073.9 & 387.66 & 150.46 & 172.78 & 0.4791 & 0.4806 & 0.4741 & 0.4787 & 581012 & 241072 & 99509 & 41102 \\
		House Price & 110.08 & 40.24 & 58.02 & 65.05 & 5178.7 & 881.2 & 726.4 & 859.4 & 0.5589 & 0.5589 & 0.5589 & 0.5587 & 2885485 & 1196674 & 496442 & 206332\\
		Stock & 262 & 121.82 & 105.98 & 127.62 & 12528.9 & 4545.9 & 1943.8 & 2169.9 & 0.5829 & 0.5828 & 0.5825 & 0.5820 & 7026593 & 2952376 & 1226666 & 508366\\
		\hline
	\end{tabular}}
	\caption{Cluster performance for IHTC with $k$-means ($t^*=2$).}
	\label{data_fkm}
\end{table}

\begin{table}[H]
	\centering
	\begin{tabular}{ c | c c c | c c c | c c c }
		 Performance & \multicolumn{3}{c}{PM 2.5} & \multicolumn{3}{c}{Credit Score} & \multicolumn{3}{c}{Black Friday} \\
		   & $m=1$ & $m=2$ & $m=3$ & $m=2$ & $m=3$ & $m=4$ & $m=1$ & $m=2$ & $m=3$ \\
		\hline
		Run Time (second) & 11.7 & 1.9 & 0.48 & 18.87 & 3.79 & 3.07 & 5.66 & 1.01 & 0.60 \\
		Memory Usage (Mb) & 3420.6 & 588.9 & 79.7 & 4799.3 & 730.98 & 21.32 & 1618.94 & 277.87 & 38.44 \\
		BSS/TSS & 0.4964 & 0.4964 & 0.4964 & 0.4746 & 0.4612 & 0.4613 & 0.3176 & 0.3024 & 0.3142 \\
		Number of Prototypes & 17281 & 7166 & 2984 & 20471 & 8456 & 3508 & 11868 & 4914 & 2017\\
		\hline
	\end{tabular}
	\caption{Cluster performance for IHTC with HAC ($t^*=2$).}
	\label{data_fhac1}
\end{table}

\begin{table}[H]
	\centering
	\begin{tabular}{ c | c c c | c c c | c c c }
		 Performance & \multicolumn{3}{c}{Covertype} &\multicolumn{3}{c}{House Price} &\multicolumn{3}{c}{Stock}\\
		 & $m=4$ & $m=5$ & $m=6$ & $m=6$ & $m=7$ & $m=8$ & $m=7$ & $m=8$ & $m=9$ \\
		\hline
		Run Time (second) & 16.34 & 15.96 & 15.72 & 63.3 & 65.1 & 64.4 & 144.24 & 144.84 & 145.12\\
		Memory Usage (Mb) & 206.53 & 211.25 & 210.1 & 940.74 & 934.97 & 933.79 & 2415.6 & 2443.9 & 2471.9\\
		BSS/TSS & 0.4124 & 0.3982 & 0.4144 & 0.5213 & 0.5017 & 0.5017 & 0.4986 & 0.4945 & 0.4993\\
		Number of Prototypes & 17015 & 7015 & 2911 & 15014 & 6268 & 2598 & 15085 & 6267 & 2603\\
		\hline
	\end{tabular}
	\caption{Cluster performance for IHTC with HAC ($t^*=2$).}
	\label{data_fhac2}
\end{table}
\end{landscape}

\section{Discussion}
\label{makereference1.6}
Many clustering methods share one ubiquitous drawback: as the size of the data becomes massive---that is, as the number of units $n$ for the data becomes large---these methods become intractable. 
Even efficient implementations of these algorithms require prohibitive computational resources under massive $n$ settings.
Recently, threshold clustering (TC) has been developed as a way of clustering units so that each cluster contains a pre-specified number of units and so that the maximum within-cluster dissimilarity is small. 
Unlike many clustering methods, TC is designed to form clustering comprised of many groups of only a few units.
Moreover, the runtime and memory requirement for TC is smaller compared to other clustering methods.

In this paper, we propose using TC as an instance selection method called iterative threshold instance selection (ITIS) to efficiently and effectively reduce the size of data.
Additionally, we propose coupling ITIS with other, more sophisticated clustering methods to obtain
method to sufficiently scale down the size of data.  
and introduced IHTC that can efficiently and effectively scale down the size of data so that more sophisticated clustering methods can be applied on the reduced data. 

Simulation results and application on real datasets show that implementing clustering methods with IHTC may decrease their runtime and memory usage without sacrificing the their performance. 
For more sophisticated clustering methods, this reduction in runtime and memory usage may be dramatic. 
Even for the standard implementation of $k$-means in R, one iteration of IHTC decreases the runtime and memory usage by more than half while maintaining clustering performance. 


\addcontentsline{toc}{chapter}{Bibliography}
\bibdata{references}
\bibliography{references}

\begin{thebibliography}{42}
\providecommand{\natexlab}[1]{#1}
\providecommand{\url}[1]{\texttt{#1}}
\expandafter\ifx\csname urlstyle\endcsname\relax
  \providecommand{\doi}[1]{doi: #1}\else
  \providecommand{\doi}{doi: \begingroup \urlstyle{rm}\Url}\fi

\bibitem[Arthur and Vassilvitskii(2007)]{arthur2007k}
David Arthur and Sergei Vassilvitskii.
\newblock k-means++: The advantages of careful seeding.
\newblock In \emph{Proceedings of the eighteenth annual ACM-SIAM symposium on
  Discrete algorithms}, pages 1027--1035. Society for Industrial and Applied
  Mathematics, 2007.

\bibitem[Blackard and Dean(1999)]{Blackard99comparativeaccuracies}
Jock~A. Blackard and Denis~J. Dean.
\newblock Comparative accuracies of artificial neural networks and discriminant
  analysis in predicting forest cover types from cartographic variables, 1999.

\bibitem[Blum and Langley(1997)]{blum1997selection}
Avrim~L Blum and Pat Langley.
\newblock Selection of relevant features and examples in machine learning.
\newblock \emph{Artificial intelligence}, 97\penalty0 (1-2):\penalty0 245--271,
  1997.

\bibitem[Brogaard et~al.(2014)Brogaard, Hendershott, and
  Riordan]{brogaard2014high}
Jonathan Brogaard, Terrence Hendershott, and Ryan Riordan.
\newblock High-frequency trading and price discovery.
\newblock \emph{The Review of Financial Studies}, 27\penalty0 (8):\penalty0
  2267--2306, 2014.

\bibitem[Carrion(2013)]{carrion2013very}
Allen Carrion.
\newblock Very fast money: High-frequency trading on the nasdaq.
\newblock \emph{Journal of Financial Markets}, 16\penalty0 (4):\penalty0
  680--711, 2013.

\bibitem[Cukier(2010)]{cukier2010data}
Kenneth Cukier.
\newblock \emph{Data, data everywhere: A special report on managing
  information}.
\newblock Economist Newspaper, 2010.

\bibitem[Dagdoug(2018)]{blackfriday}
Mehdi Dagdoug.
\newblock Black friday: A study of sales trough consumer behaviours, 2018.
\newblock URL \url{https://www.kaggle.com/mehdidag/black-friday}.

\bibitem[Ester et~al.(1996)Ester, Kriegel, Sander, Xu,
  et~al.]{ester1996density}
Martin Ester, Hans-Peter Kriegel, J{\"o}rg Sander, Xiaowei Xu, et~al.
\newblock A density-based algorithm for discovering clusters in large spatial
  databases with noise.
\newblock In \emph{Kdd}, volume~96, pages 226--231, 1996.

\bibitem[Firdaus and Uddin(2015)]{firdaus2015survey}
Sabhia Firdaus and Md~Ashraf Uddin.
\newblock A survey on clustering algorithms and complexity analysis.
\newblock \emph{International Journal of Computer Science Issues (IJCSI)},
  12\penalty0 (2):\penalty0 62, 2015.

\bibitem[Fr{\"a}nti and Kivij{\"a}rvi(2000)]{franti2000randomised}
Pasi Fr{\"a}nti and Juha Kivij{\"a}rvi.
\newblock Randomised local search algorithm for the clustering problem.
\newblock \emph{Pattern Analysis \& Applications}, 3\penalty0 (4):\penalty0
  358--369, 2000.

\bibitem[Fr{\"a}nti et~al.(1997)Fr{\"a}nti, Kivij{\"a}rvi, Kaukoranta, and
  Nevalainen]{franti1997genetic}
Pasi Fr{\"a}nti, Juha Kivij{\"a}rvi, Timo Kaukoranta, and Olli Nevalainen.
\newblock Genetic algorithms for large-scale clustering problems.
\newblock \emph{The Computer Journal}, 40\penalty0 (9):\penalty0 547--554,
  1997.

\bibitem[Friedman et~al.(2001)Friedman, Hastie, and
  Tibshirani]{friedman2001elements}
Jerome Friedman, Trevor Hastie, and Robert Tibshirani.
\newblock \emph{The elements of statistical learning}, volume~1.
\newblock Springer series in statistics Springer, Berlin, 2001.

\bibitem[Friedman et~al.(1976)Friedman, Bentley, and
  Finkel]{friedman1976algorithm}
Jerome~H Friedman, Jon~Louis Bentley, and Raphael~Ari Finkel.
\newblock An algorithm for finding best matches in logarithmic time.
\newblock \emph{ACM Trans. Math. Software}, 3\penalty0 (SLAC-PUB-1549-REV.
  2):\penalty0 209--226, 1976.

\bibitem[Frnti et~al.(1998)Frnti, Kivijrvi, and Nevalainen]{frnti1998tabu}
P~Frnti, J~Kivijrvi, and OLLI Nevalainen.
\newblock Tabu search algorithm for codebook generation in vector quantization.
\newblock \emph{Pattern Recognition}, 31\penalty0 (8):\penalty0 1139--1148,
  1998.

\bibitem[Hartigan and Wong(1979)]{hartigan1979algorithm}
John~A Hartigan and Manchek~A Wong.
\newblock Algorithm as 136: A k-means clustering algorithm.
\newblock \emph{Journal of the Royal Statistical Society. Series C (Applied
  Statistics)}, 28\penalty0 (1):\penalty0 100--108, 1979.

\bibitem[Hastie et~al.(2009)Hastie, Tibshirani, and
  Friedman]{hastie2009unsupervised}
Trevor Hastie, Robert Tibshirani, and Jerome Friedman.
\newblock Unsupervised learning.
\newblock In \emph{The elements of statistical learning}, pages 485--585.
  Springer, 2009.

\bibitem[Havera(2017)]{pm25}
David Havera.
\newblock Beijing pm2.5 data data set, 2017.
\newblock URL
  \url{https://www.kaggle.com/djhavera/beijing-pm25-data-data-set/activity}.

\bibitem[Higgins et~al.(2016)Higgins, S{\"a}vje, and
  Sekhon]{higgins2016improving}
Michael~J Higgins, Fredrik S{\"a}vje, and Jasjeet~S Sekhon.
\newblock Improving massive experiments with threshold blocking.
\newblock \emph{Proceedings of the National Academy of Sciences}, 113\penalty0
  (27):\penalty0 7369--7376, 2016.

\bibitem[Hochbaum and Shmoys(1986)]{hochbaum1986unified}
Dorit~S Hochbaum and David~B Shmoys.
\newblock A unified approach to approximation algorithms for bottleneck
  problems.
\newblock \emph{Journal of the ACM (JACM)}, 33\penalty0 (3):\penalty0 533--550,
  1986.

\bibitem[Hotelling(1933)]{hotelling1933analysis}
Harold Hotelling.
\newblock Analysis of a complex of statistical variables into principal
  components.
\newblock \emph{Journal of educational psychology}, 24\penalty0 (6):\penalty0
  417, 1933.

\bibitem[Jain et~al.(1999)Jain, Murty, and Flynn]{jain1999data}
Anil~K Jain, M~Narasimha Murty, and Patrick~J Flynn.
\newblock Data clustering: a review.
\newblock \emph{ACM computing surveys (CSUR)}, 31\penalty0 (3):\penalty0
  264--323, 1999.

\bibitem[Jordan and Mitchell(2015)]{jordan2015machine}
Michael~I Jordan and Tom~M Mitchell.
\newblock Machine learning: Trends, perspectives, and prospects.
\newblock \emph{Science}, 349\penalty0 (6245):\penalty0 255--260, 2015.

\bibitem[Kaggle(2011)]{score}
Kaggle.
\newblock Give me some credit.
\newblock Website, 2011.
\newblock URL \url{https://www.kaggle.com/c/GiveMeSomeCredit/data}.
\newblock last checked: 12.08.2018.

\bibitem[Knuth(1998)]{knuth1998art}
Donald~Ervin Knuth.
\newblock \emph{The art of computer programming: sorting and searching},
  volume~3.
\newblock Pearson Education, 1998.

\bibitem[Kurita(1991)]{kurita1991efficient}
Takio Kurita.
\newblock An efficient agglomerative clustering algorithm using a heap.
\newblock \emph{Pattern Recognition}, 24\penalty0 (3):\penalty0 205--209, 1991.

\bibitem[Leyva et~al.(2015)Leyva, Gonz{\'a}lez, and P{\'e}rez]{leyva2015three}
Enrique Leyva, Antonio Gonz{\'a}lez, and Ra{\'u}l P{\'e}rez.
\newblock Three new instance selection methods based on local sets: A
  comparative study with several approaches from a bi-objective perspective.
\newblock \emph{Pattern Recognition}, 48\penalty0 (4):\penalty0 1523--1537,
  2015.

\bibitem[Liu and Motoda(1998)]{liu1998feature}
Huan Liu and Hiroshi Motoda.
\newblock \emph{Feature extraction, construction and selection: A data mining
  perspective}, volume 453.
\newblock Springer Science \& Business Media, 1998.

\bibitem[Liu and Motoda(2002)]{liu2002issues}
Huan Liu and Hiroshi Motoda.
\newblock On issues of instance selection.
\newblock \emph{Data Mining and Knowledge Discovery}, 6\penalty0 (2):\penalty0
  115--130, 2002.

\bibitem[Lloyd(1982)]{lloyd1982least}
Stuart Lloyd.
\newblock Least squares quantization in pcm.
\newblock \emph{IEEE transactions on information theory}, 28\penalty0
  (2):\penalty0 129--137, 1982.

\bibitem[Olvera-L{\'o}pez et~al.(2007)Olvera-L{\'o}pez, Carrasco-Ochoa, and
  Mart{\'\i}nez-Trinidad]{olvera2007object}
J~Arturo Olvera-L{\'o}pez, J~Ariel Carrasco-Ochoa, and J~Francisco
  Mart{\'\i}nez-Trinidad.
\newblock Object selection based on clustering and border objects.
\newblock In \emph{Computer Recognition Systems 2}, pages 27--34. Springer,
  2007.

\bibitem[Olvera-L{\'o}pez et~al.(2008)Olvera-L{\'o}pez, Carrasco-Ochoa, and
  Mart{\'\i}nez-Trinidad]{olvera2008prototype}
J~Arturo Olvera-L{\'o}pez, J~Ariel Carrasco-Ochoa, and J~Fco
  Mart{\'\i}nez-Trinidad.
\newblock Prototype selection via prototype relevance.
\newblock In \emph{Iberoamerican Congress on Pattern Recognition}, pages
  153--160. Springer, 2008.

\bibitem[Olvera-L{\'o}pez et~al.(2010)Olvera-L{\'o}pez, Carrasco-Ochoa,
  Mart{\'\i}nez-Trinidad, and Kittler]{olvera2010review}
J~Arturo Olvera-L{\'o}pez, J~Ariel Carrasco-Ochoa, J~Francisco
  Mart{\'\i}nez-Trinidad, and Josef Kittler.
\newblock A review of instance selection methods.
\newblock \emph{Artificial Intelligence Review}, 34\penalty0 (2):\penalty0
  133--143, 2010.

\bibitem[Plasencia-Cala{\~n}a et~al.(2014)Plasencia-Cala{\~n}a, Orozco-Alzate,
  M{\'e}ndez-V{\'a}zquez, Garc{\'\i}a-Reyes, and Duin]{plasencia2014towards}
Yenisel Plasencia-Cala{\~n}a, Mauricio Orozco-Alzate, Heydi
  M{\'e}ndez-V{\'a}zquez, Edel Garc{\'\i}a-Reyes, and Robert~PW Duin.
\newblock Towards scalable prototype selection by genetic algorithms with fast
  criteria.
\newblock In \emph{Joint IAPR International Workshops on Statistical Techniques
  in Pattern Recognition (SPR) and Structural and Syntactic Pattern Recognition
  (SSPR)}, pages 343--352. Springer, 2014.

\bibitem[Pooja(2013)]{pooja2013comparative}
Saroj~Ratnoo Pooja.
\newblock A comparative study of instance reduction techniques.
\newblock In \emph{Proceedings of 2nd International Conference on Emerging
  Trends in Engineering and Management, ICETEM}. Citeseer, 2013.

\bibitem[Raicharoen and Lursinsap(2005)]{raicharoen2005divide}
Thanapant Raicharoen and Chidchanok Lursinsap.
\newblock A divide-and-conquer approach to the pairwise opposite class-nearest
  neighbor (poc-nn) algorithm.
\newblock \emph{Pattern recognition letters}, 26\penalty0 (10):\penalty0
  1554--1567, 2005.

\bibitem[Riquelme et~al.(2003)Riquelme, Aguilar-Ruiz, and
  Toro]{riquelme2003finding}
Jos{\'e}~C Riquelme, Jes{\'u}s~S Aguilar-Ruiz, and Miguel Toro.
\newblock Finding representative patterns with ordered projections.
\newblock \emph{Pattern Recognition}, 36\penalty0 (4):\penalty0 1009--1018,
  2003.

\bibitem[Savje et~al.(2017)Savje, Higgins, and Sekhon]{scclust}
Fredrik Savje, Michael Higgins, and Jasjeet Sekhon.
\newblock \emph{scclust: Size-Constrained Clustering}, 2017.
\newblock URL \url{https://CRAN.R-project.org/package=scclust}.
\newblock R package version 0.1.1.

\bibitem[Sullivan(2015)]{sullivan2015google}
Danny Sullivan.
\newblock Google still doing at least 1 trillion searches per year.
\newblock \emph{Retrieved September}, 29:\penalty0 2015, 2015.

\bibitem[University(2018)]{BEOCAT}
Kansas~State University.
\newblock Acknowledging use of beocat resources and/or personnel in
  publications, 2018.
\newblock URL \url{https://support.beocat.ksu.edu/BeocatDocs/index.php/Policy}.

\bibitem[Vaidya(1989)]{vaidya1989ano}
Pravin~M Vaidya.
\newblock An $o(n \log n)$ algorithm for the all-nearest-neighbors problem.
\newblock \emph{Discrete \& Computational Geometry}, 4\penalty0 (2):\penalty0
  101--115, 1989.

\bibitem[Ward~Jr(1963)]{ward1963hierarchical}
Joe~H Ward~Jr.
\newblock Hierarchical grouping to optimize an objective function.
\newblock \emph{Journal of the American statistical association}, 58\penalty0
  (301):\penalty0 236--244, 1963.

\bibitem[Zillow(2017)]{zillow}
Zillow.
\newblock Zillow prize: Zillow’s home value prediction (zestimate), 2017.
\newblock URL \url{https://www.kaggle.com/c/zillow-prize-1}.

\end{thebibliography}


\cleardoublepage
\appendix
\section{Cluster Performance for IHTC with Varying Threshold Size}
\label{AppendixA}
\begin{figure}[H]
\centering
	\minipage{0.5\textwidth}	\includegraphics[width=\linewidth,page=3]{simulation/fkm_avg_merge.pdf}
	\label{fkm_time}
	\endminipage
	\minipage{0.5\textwidth}
\includegraphics[width=\linewidth,page=1]{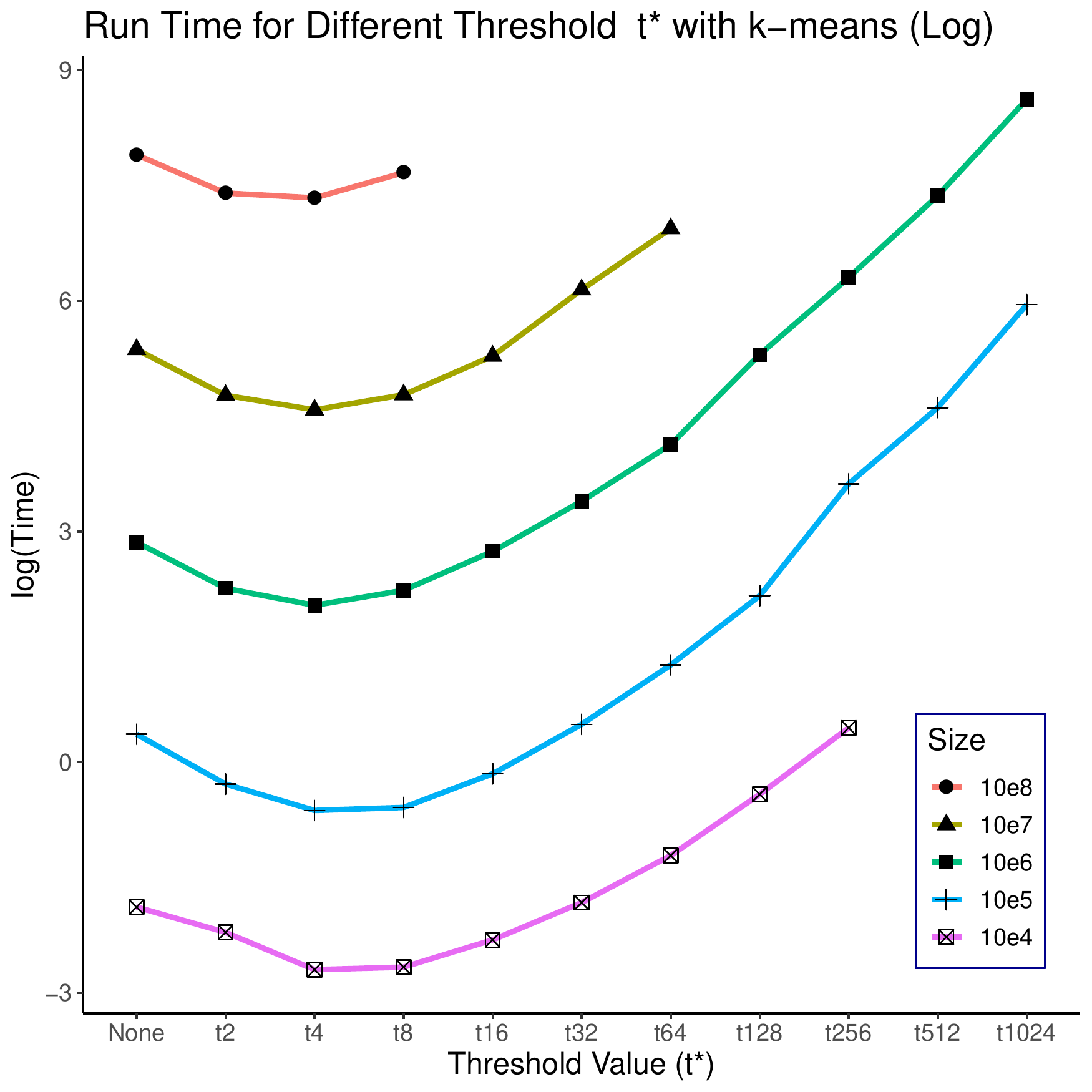}
	\label{fkm_logtime}
	\endminipage\hfill
	\minipage{0.5\textwidth}	\includegraphics[width=\linewidth,page=2]{simulation/fkm_avg_merge.pdf}
	\label{fkm_mem}
	\endminipage
	\minipage{0.5\textwidth}
\includegraphics[width=\linewidth,page=5]{simulation/fkm_avg_merge.pdf}
	\label{fkm_logmem}
	\endminipage
	\caption{Run time and memory usage of different t* with $k$-means ($k=3,m=1$).}
	\label{fig_fkm}
\end{figure}

\begin{figure}[H]
\centering
	\minipage{0.5\textwidth}	\includegraphics[width=\linewidth,page=3]{simulation/fhac_merge.pdf}
	\label{fhac_time}
	\endminipage
	\minipage{0.5\textwidth}
\includegraphics[width=\linewidth,page=1]{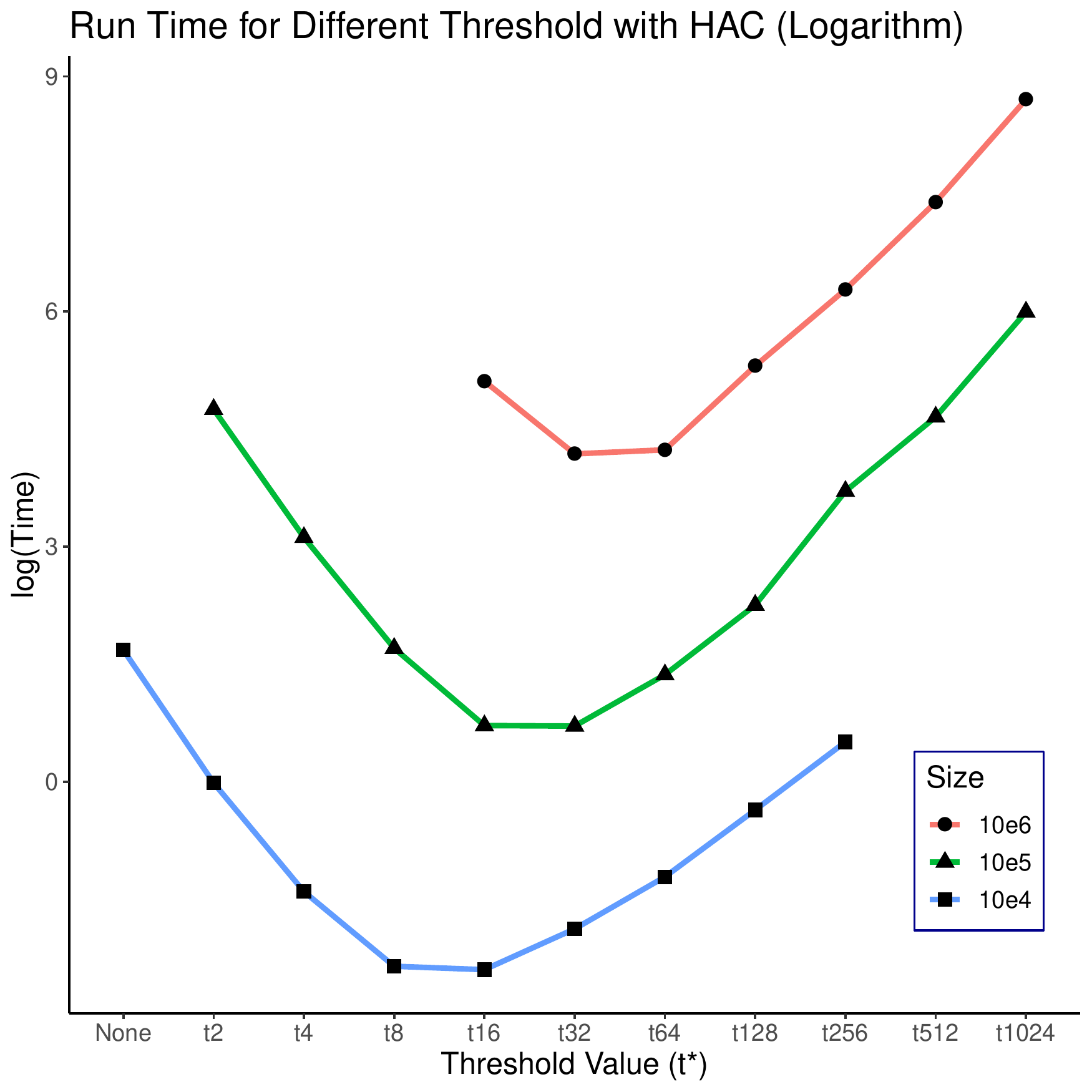}
	\label{fhac_logtime}
	\endminipage\hfill
	\minipage{0.5\textwidth}	\includegraphics[width=\linewidth,page=2]{simulation/fhac_merge.pdf}
	\label{fhac_mem}
	\endminipage
	\minipage{0.5\textwidth}
\includegraphics[width=\linewidth,page=5]{simulation/fhac_merge.pdf}
	\label{fhac_logmem}
	\endminipage
	\caption{Runtime and memory usage of for different threshold $t^*$ with HAC ($m=1$).}
	\label{fig_htc_hac}
\end{figure}

\begin{figure}[H]
\centering
    \minipage{0.5\textwidth}
	\includegraphics[width=\linewidth,page=4]{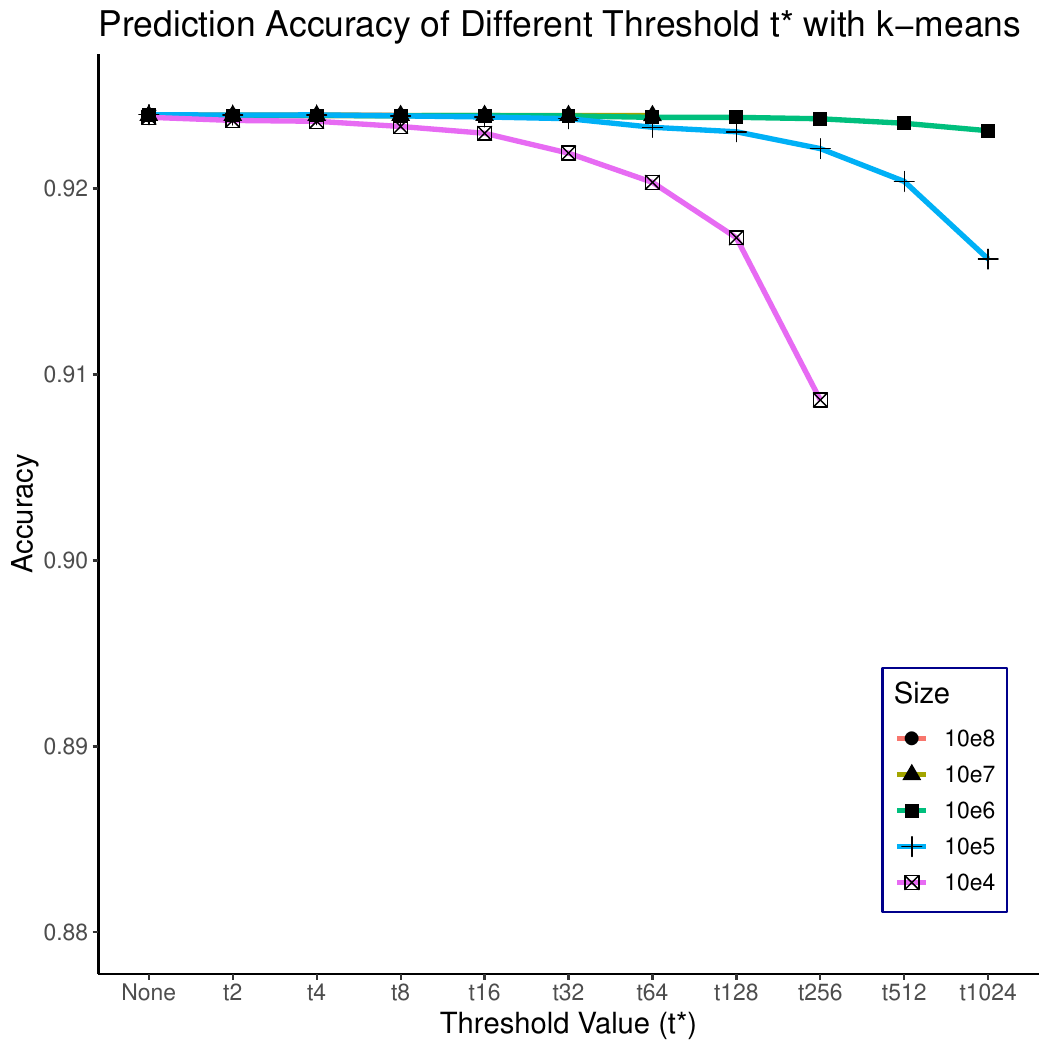}
	\label{fig_htc_kmaccu}
	\endminipage
	\minipage{0.5\textwidth}
	\includegraphics[width=\linewidth,page=4]{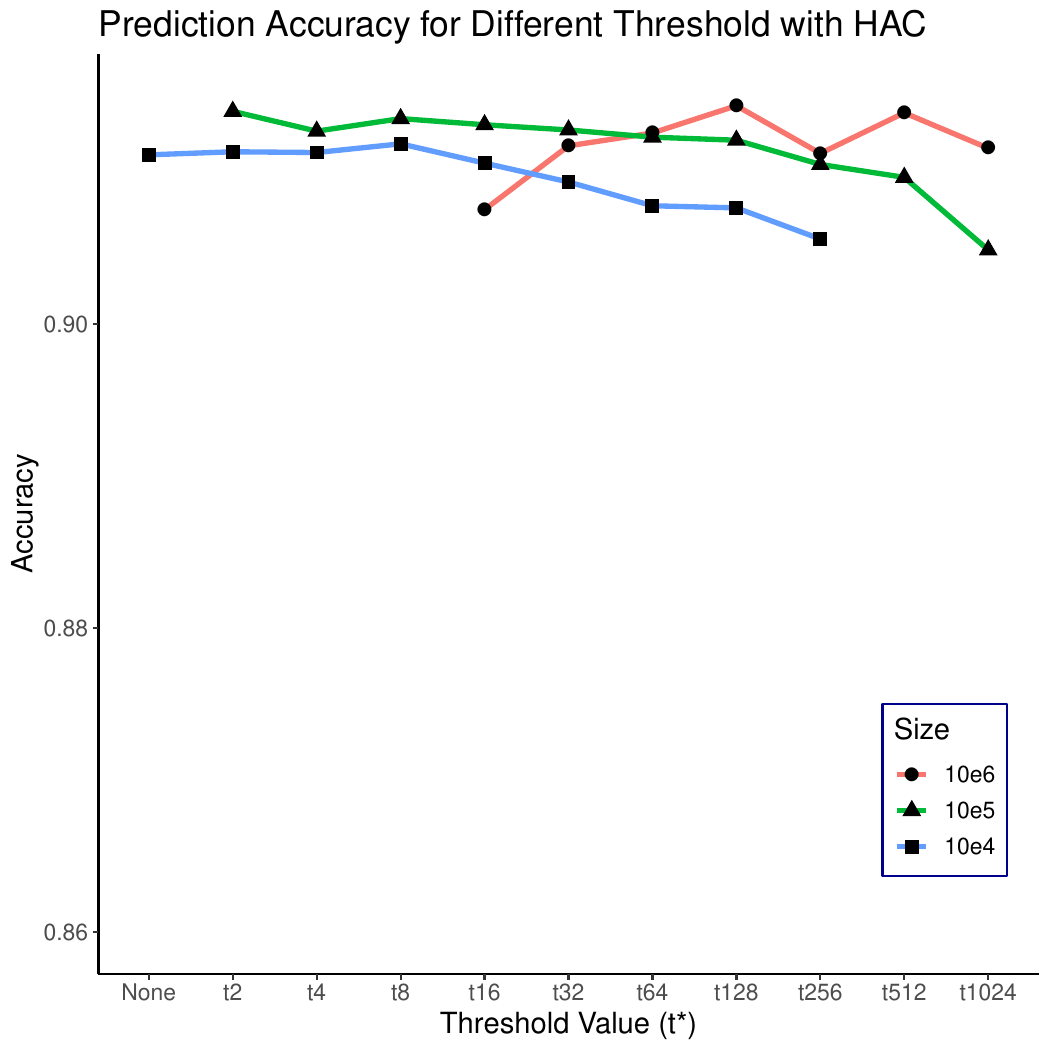}
	\label{fig_htc_hacaccu}
	\endminipage
	\caption{Prediction accuracy of for different threshold $t^*$ ($m=1$).}
	\label{fig_htc_accu}
\end{figure}

We compare the cluster performance for IHTC with varying threshold $t^*$ across different data size. 
For this example, we generate data using the multivariate Gaussian model in Section~\ref{makereference1.4}.
We set $k=3$, $n$ is between $10^4$ and $10^8$, and perform one iteration of IHTC ($m = 1$). 
We analyze performance across different thresholds: $t^*=2,4,8,16,32,64,128,256,512$ and $1024$.
The runtime, memory usage and prediction accuracy for IHTC with $k$-means can be found in Figures~\ref{fig_fkm} and~\ref{fig_htc_accu}, and Table~\ref{t_fkm}. 
Figures~\ref{fig_htc_hac} and~\ref{fig_htc_accu}, and
Table~\ref{t_fhac} present the cluster performance for IHTC with HAC.

We find, that when the threshold $t^*$ is small, pre-processing the data with IHTC decreases runtime and memory usage compared to without pre-processing, and prediction accuracy fluctuates within a narrow range. 
When $t^*$ is large, the runtime for our method takes longer than the runtime without pre-processing. 
In general, across all data sizes, the runtime initially decreases before steadily increasing as $t^*$ increases. 
On the other hand, despite increasing the threshold $t^*$, the memory usage for IHTC with $k$-means or HAC is continually reducing. 
Additionally, the prediction accuracy decreases slightly with larger values of $t^*$. 

\newpage

\begin{landscape}
\begin{table}[H]
	\centering
	\resizebox{\linewidth}{!}{%
	\begin{tabular}{ c | c c c c c | c c c c c | c c c c c }
		 & \multicolumn{5}{c}{Run Time (second)} & \multicolumn{5}{c}{Memory Usage (Mb)} &\multicolumn{5}{c}{Accuracy}\\
		 $t^*$ & $10^4$ & $10^5$ & $10^6$ & $10^7$ & $10^8$ & $10^4$ & $10^5$ & $10^6$ & $10^7$ & $10^8$ & $10^4$ & $10^5$ & $10^6$ & $10^7$ & $10^8$\\
		\hline
		None   & 0.152 &  1.44 &  17.5 & 214 & 2697 & 20.37 & 235.36 & 2538 & 27468 & 278617  &   0.9238 & 0.9240 & 0.9239 & 0.9239 & 0.9239\\
		2 & 0.109 &  0.75  & 9.6 & 118 & 1640 &  9.50  & 96.73 & 1014 & 11085 & 110919 &  0.9237 & 0.9240 & 0.9239 & 0.9239 & 0.9239\\
		4 & 0.067  & 0.53  &  7.7 &  98 & 1540 &  4.33  & 48.17 &  525 &  5898  & 59288 &   0.9236 & 0.9240 & 0.9239 & 0.9239 & 0.9239\\
		8 & 0.070  & 0.56 &  9.4 & 119 & 2137 &  1.65  & 24.50 &  274 &  3293 &  33316 &   0.9233 & 0.9239 & 0.9239 & 0.9239 & 0.9239\\
		16 & 0.099 & 0.86 &  15.5 & 197 &  - &  0.52  & 11.97 &  149  & 1975  &    - &   0.9230 & 0.9238 & 0.9239 & 0.9239   &   -\\
		32 & 0.161 &  1.64 &  29.8 & 467 &  - &  0.20   & 5.34 &   90 &  1314  &    - &   0.9219 & 0.9238 & 0.9239 & 0.9239  &    -\\
		64 &  0.297 &  3.54 &  62.3 & 1032 &  - &  0.10  &  2.21 &   59 &   999  &    - &   0.9203 & 0.9233 & 0.9238 & 0.9239   &   -\\
		128 & 0.658  & 8.72 & 200.8  & -  & - &  0.04  &  0.98 &   41  &     -  &    - &   0.9174 & 0.9231 & 0.9238  &    -  &    -\\
		256 & 1.565 & 37.38 & 546.4 &  -  & - &  0.03 &   0.58 &   34  &     -  &    - &   0.9086 & 0.9221 & 0.9238  &    -  &    -\\
		512 & - & 100.56 & 1585 & - &  - &  - &   0.45 &   27  &     -  &    - &   - & 0.9204 & 0.9235  &    -  &    -\\
		1024  &   - & 384.17 & 5541 & - &  - &    -  &  0.39  &  26  &     -   &   - &     - & 0.9162 & 0.9231  &    -  &    -\\
		\hline
	\end{tabular}}
	\caption{Cluster performance for iterate once with $k$-means ($k=3,m=1$).}
	\label{t_fkm}
\end{table}

\begin{table}[H]
	\centering
	\begin{tabular}{ c | c c c | c c c| c c c }
		& \multicolumn{3}{c}{Run Time (s)}&\multicolumn{3}{c}{Memory Usage (Mb)}&\multicolumn{3}{c}{Accuracy} \\
		$t^*$ & $10^4$ & $10^5$ & $10^6$ & $10^4$ & $10^5$ & $10^6$ & $10^4$ & $10^5$ & $10^6$\\
		\hline
		Null  &   5.366  &     -   &    - & 764.5 &    -   &     - & 0.9111   &   -  &    -\\
		2  &   0.984 & 115.9   &    - & 184.8 & 20196   &     - & 0.9113 & 0.9140   &   -\\
		4  &   0.246  & 22.59   &    - &  24.84 & 4992   &     - & 0.9113 & 0.9127  &    -\\
		8  &   0.095  &  5.491   &    - &  12.46 & 1230   &     - & 0.9118 & 0.9135  &    -\\
		16 &   0.091  &  2.047 &  165.8 &   1.720 & 298.4 & 29701 & 0.9106 & 0.9131 & 0.9075\\
		32 &   0.153  &  2.035 &   65.59 &   0.066 & 73.78 &  7207 & 0.9093 & 0.9127 & 0.9117\\
		64 &   0.297  &  3.919 &   69.05 &   0.018 & 19.17 &  1813 & 0.9078 & 0.9123 & 0.9125\\
		128&   0.697  &  9.522 &  201.3 &   0.005 & 3.517 &   466.2 & 0.9076 & 0.9121 & 0.9143\\
		256 &  1.662  & 40.73 &  534.3 &   0.004 & 0.561  &  130.2 & 0.9056 & 0.9105 & 0.9112\\
		512 &  - & 105.2 & 1630  &  - & 0.325  &   47.29 & - & 0.9096 & 0.9139\\
		1024  &     - & 401.4 &  6042  &      - & 0.241  &   26.97  &    - & 0.9049 & 0.9115\\
		\hline
	\end{tabular}
	\caption{Simulation result for different threshold $t^*$ with HAC ($m=1$).}
	\label{t_fhac}
\end{table}
\end{landscape}

\newpage
\section{Experiment for IHTC with DBSCAN}
\label{AppendixB}

Table~\ref{data_dbscan} shows the result on the four datasets with the fewest instances.
The parameters $\epsilon$ and $Minpts$ are decided by subsample of size 1000 of each dataset with a 10-fold cross-validation method. 
Comparing the performance for DBSCAN with and without IHTC, we find that IHTC with DBSCAN has shorter runtime but higher memory usage than DBSCAN itself. 
Total sum of squares (TSS) is equal to between-cluster sum of squares (BSS) plus within-cluster sum of squares. 
Higher ratio of BSS and TSS indicates the clusters are more compact. 
We found that the ratio of BSS and TSS is higher when applying IHTC, which shows our method has comparable clustering performance.


\begin{table}[H]
	\centering
	\resizebox{\textwidth }{!}{%
	\begin{tabular}{ c | c c c | c c c | c c c }
		 Name& \multicolumn{3}{c}{Run Time (second)} & \multicolumn{3}{c}{Memory Usage (Mb)} &\multicolumn{3}{c}{BSS/TSS} \\
		  & $m=0$ & $m=1$ & $m=2$ & $m=0$ & $m=1$ & $m=2$ & $m=0$ & $m=1$ & $m=2$ \\
		\hline
		PM 2.5 & 3.96 & 0.9 & 0.26 & 0.8 & 0.4 & 0.2 & 0.5036 & 0.5627 & 0.5336\\
		Credit Score & 25.78 & 4.16 & 2.66 & 3.3 & 1.2 & 38 & 0.4731 & 0.6015 & 0.6441\\
		Black Fridy & 9.26 & 0.64 & 0.56 & 13.5 & 62.1 & 66.2 & 0.3103 & 0.9657 & 0.9985\\
		Covertype & 233.9 & 223.8 & 231.4 & 13.3 & 15.6 & 15.6 & 0.1785 & 0.1785 & 0.1785\\
		\hline
	\end{tabular}}
	\caption{Cluster performance for IHTC with DBSCAN ($t^*=2$).}
	\label{data_dbscan}
\end{table}

\newpage
\section{Graph theory definitions}
\label{sec:grthrdef}
Let $G = (V,E)$ denote an undirected graph. \begin{definition}\label{def_adjacent}
    Vertices $i$ and $j$ are \textit{adjacent} in $G$ if the edge $ij \in E$.
\end{definition}
  
\begin{definition}\label{def_independent}
	A set of vertices $I \subseteq V$ is \textit{independent in $G$} if no vertices in the set are adjacent to each other.
    That is, for all $i,j \in I$,  $ij \notin E$.
\end{definition}
  
\begin{definition}\label{def_maximal}
    An independent set of vertices $I$ in $G$ is \textit{maximal} if, for any additional vertex $i \in V$, the set $i\cup I$ is not independent. 
    That is, for all $i \in  V \setminus I$, there exists $j\in I$ such that $ij \in E$.
\end{definition}
  

\begin{definition}\label{def_walk}
    For $i,j \in V$, a \textit{walk from $i$ to $j$ of length $m$ in $G$} is a sequence of $m+1$ vertices
    $(i = i_0, i_1, \ldots, i_m = j)$ such that the edge $i_{\ell-1}i_{\ell} \in E$.  
\end{definition}
\noindent Note that, if $(i = i_0, i_1, i_2, \cdots, i_{m} = j)$ is a walk of length $m$ from $i$ to $j$ and the edge weights of $G$ satisfy the triangle inequality~\eqref{eq:triangleineq}, then the weight $w_{ij}$ satisfies the inequality:    
	\begin{equation}
	w_{ij} \leq m\max_{\ell}(w_{i_{\ell -1} i_{\ell}}).
	\end{equation}
    
\begin{definition}\label{def_power}
	The \textit{$d^\text{th}$ power} of $G$, denoted $G^d = (V, E^d)$, is a graph such that an edge $ij \in E^d$ if and only if there is a walk from $i$ to $j$ of length at most $d$ in $G$.
\end{definition}


\begin{definition}\label{def_nng}
The \textit{$k$-nearest-neighbors subgraph of $G$} is a subgraph $NG_k =(V, NE_k)$ where an edge $ij \in NE_{k}$ if and only if $j$ is one of the $k$ closest vertices to $i$ or $i$ is one of the $k$ closest vertices to $j$.
Rigorously, for a vertex $i$, let $i_{(\ell)}$ denote the vertex that corresponds to the $\ell^\text{th}$ smallest value of $w_{ij},  ij \in E$, where ties are broken arbitrarily: $w_{ii_{(1)}} \leq w_{ii_{(2)}} \leq \cdots \leq w_{ii_{(m)}}$. 
Then
\begin{equation}
NE_k \equiv \left\{ij\in E :j \in (i_{(\ell)})_{\ell = 1}^k \text{ or } i \in (j_{(\ell)})_{\ell = 1}^k\right\}.
\end{equation}
\end{definition}

\end{document}